\documentclass[10pt,twocolumn,letterpaper]{article}

\usepackage{cvpr}
\usepackage{times}
\usepackage{epsfig}
\usepackage{graphicx}
\usepackage{amsmath}
\usepackage{amssymb}

\usepackage{algorithm}
\usepackage{algorithmic}
\usepackage{subfig}

\usepackage[pagebackref=true,breaklinks=true,letterpaper=true,colorlinks,bookmarks=false]{hyperref}

\cvprfinalcopy 

\def\cvprPaperID{672} 

\newcommand{\bm}{\boldsymbol}
\newcommand{\ten}[1]{ \boldsymbol{\mathcal #1}}
\newcommand{\bbR}[1]{\mathbb{R}^{#1}}

\usepackage{lipsum}
\newcommand\blfootnote[1]{%
  \begingroup
  \renewcommand\thefootnote{}\footnote{#1}%
  \addtocounter{footnote}{-1}%
  \endgroup
}

\ifcvprfinal\fi

\begin{document}
\title{Missing Slice Recovery for Tensors Using a Low-rank Model in Embedded Space}

\author{Tatsuya Yokota$^1$ \quad Burak Erem$^2$ \quad Seyhmus Guler$^2$ \quad Simon K. Warfield$^2$ \quad Hidekata Hontani$^1$ \\
$^1$Nagoya Institute of Technology, Gokiso, Showa, Nagoya, Japan\\
$^2$Boston Children's Hospital, 300 Longwood Ave., Boston, MA, United States\\
{\tt\footnotesize \{t.yokota, hontani\}@nitech.ac.jp, \{burak.erem, seyhmus.guler, simon.warfield\}@childrens.harvard.edu}
}

\maketitle

\begin{abstract}
  Let us consider a case where all of the elements in some continuous slices are missing in tensor data.
  In this case, the nuclear-norm and total variation regularization methods usually fail to recover the missing elements.
  The key problem is capturing some delay/shift-invariant structure.
  In this study, we consider a low-rank model in an embedded space of a tensor.
  For this purpose, we extend a delay embedding for a time series to a ``multi-way delay-embedding transform'' for a tensor, which takes a given incomplete tensor as the input and outputs a higher-order incomplete Hankel tensor.
  The higher-order tensor is then recovered by Tucker-based low-rank tensor factorization.
  Finally, an estimated tensor can be obtained by using the inverse multi-way delay embedding transform of the recovered higher-order tensor.
  Our experiments showed that the proposed method successfully recovered missing slices for some color images and functional magnetic resonance images.
\end{abstract}


\section{Introduction}
\blfootnote{This work was supported in part by JSPS Grant-in-Aid for Scientific
Research on Innovative Areas (Multidisciplinary Computational
Anatomy): JSPS KAKENHI Grant Number 26108003 and 15K16067.}
Matrix/tensor completion is a technique for recovering the missing elements in incomplete data and it has become a very important method in recent years \cite{candes2009exact,candes2010power,candes2010matrix,gandy2011tensor,liu2013tensor,shi2013low,chen2014simultaneous,zhao2015bayesian,yokota2016smooth}.
In general, completion is an ill-posed problem without any assumptions.
However, if we have useful prior knowledge or assumptions regarding the data structure, completion can be treated as a well-posed optimization problem, such as convex optimization.
The assumption of the structure is also referred to as a ``model.''

The methods for modeling matrices/tensors can be categorized into two classes.
In the first class, the methods directly represent data with the matrices/tensors themselves and some structures of the matrices/tensors are assumed, such as low-rank \cite{candes2009exact,candes2010power,candes2010matrix,gandy2011tensor,liu2013tensor} and smooth properties \cite{zhu2008efficient,guo2015generalized}.

\begin{figure}[t]
  \centering
  \includegraphics[width=0.49\textwidth]{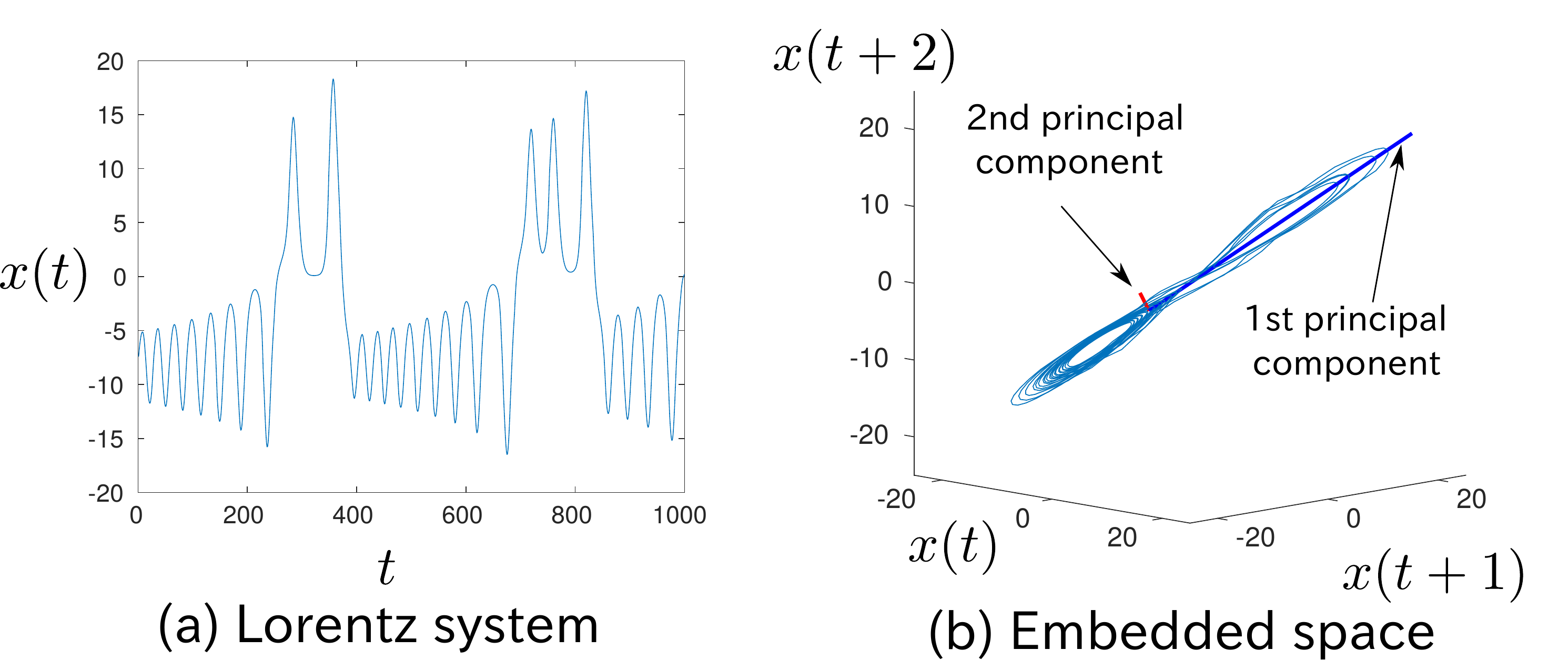}
  \caption{Lorentz system and its delay embedded space: A time series signal $x(t)$ can be embedded into a three-dimensional space with individual axes of $x(t)$, $x(t+1)$, and $x(t+2)$. Clearly, most of the points are located on some hyper-plane (i.e., low-rank) in the embedded space.}\label{fig:Lorentz}
\end{figure}

\begin{figure}[t]
  \centering
  \includegraphics[width=0.49\textwidth]{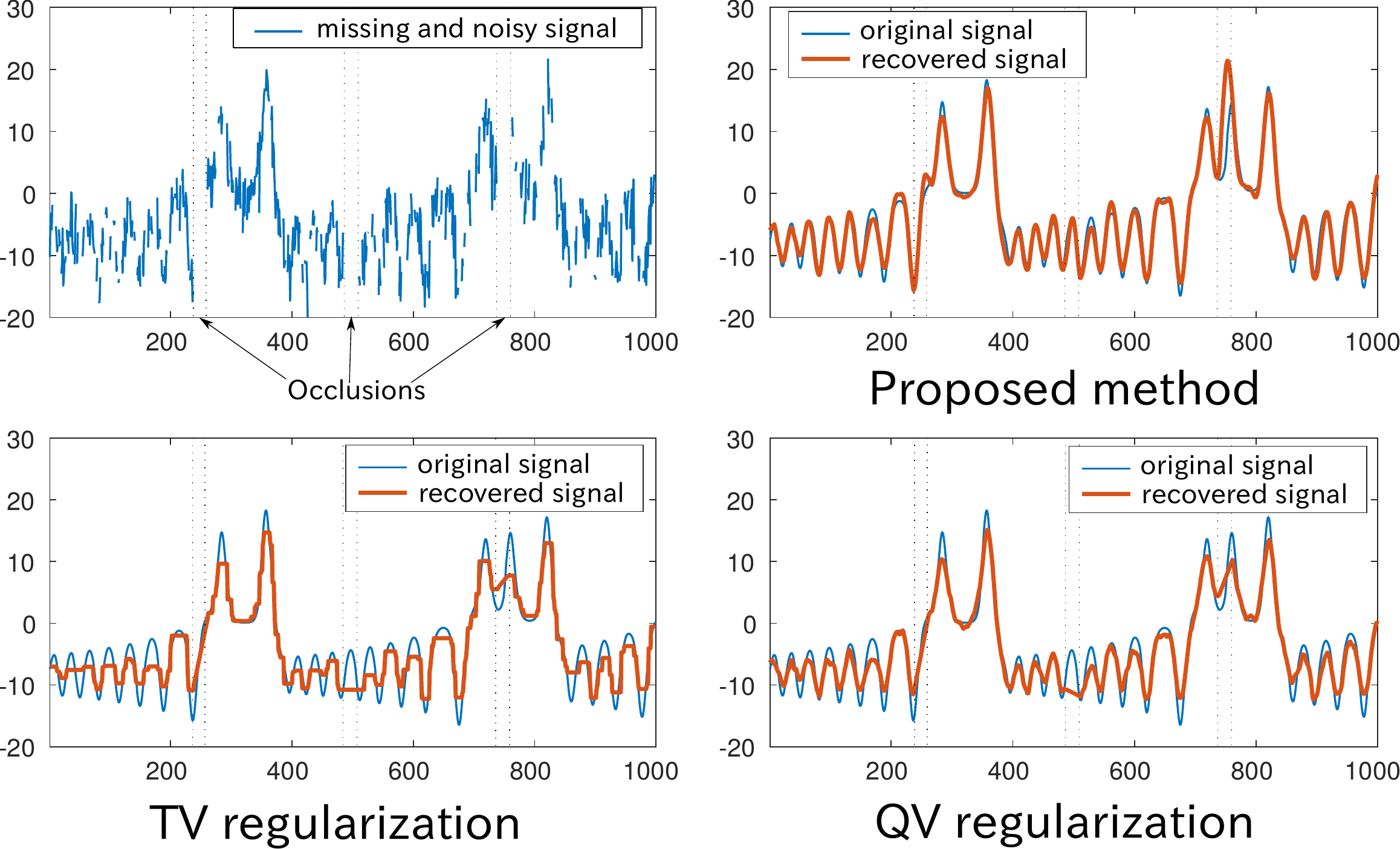}
  \caption{Recovering a dynamical signal using the proposed method ($\tau=50$), TV regularization, and QV regularization.}\label{fig:toy_comp}
\end{figure}

By contrast, the methods in the second class ``embed'' the data into a high-dimensional feature space and it is assumed that the data can be represented by low-rank or a smooth manifold in the embedded space \cite{lorenz1963deterministic,van1991subspace,ding2007rank,markovsky2008structured} (see Figure~\ref{fig:Lorentz}).
Typically, a time series is represented by a ``Hankel matrix'' (see Section~\ref{sec:Hankelization}) and its low-rank property has been employed widely for modeling a {\em linear time-invariant system} of signals \cite{van1991subspace,markovsky2008structured}.
For example, Li \etal \cite{li1997parameter} proposed a method for modeling damped sinusoidal signals based on a low-rank Hankel approximation.
Ding \etal \cite{ding2007rank} proposed the use of rank minimization of a Hankel matrix for the video inpainting problem by assuming an autoregressive moving average model.
Figure~\ref{fig:toy_comp} shows an example of occlusion recovery for a noisy time series, which
indicates that total variation (TV) and quadratic variation (QV) regularization methods reconstruct a flat estimator, whereas minimization of the Hankel matrix (our proposed method) successfully reconstructs the signal.

In the proposed method, the incomplete input data are not represented as a Hankel matrix, but instead they are represented as a ``higher order Hankel tensor'' via multi-way embedding with delay/shift along the time/space axes, and we solve the low-rank tensor completion problem in the embedded space.
The minimization of the rank of a matrix/tensor is NP-hard \cite{gillis2011low} and the problem is often relaxed to nuclear-norm minimization \cite{recht2010guaranteed}.
A disadvantage of the relaxation to nuclear norm minimization is that it decreases the rank of the resultant matrix/tensor as well as the total component values in the matrix/tensor.
In particular, nuclear norm minimization often obtains ``dark'' signals in denoising tasks.
Thus, we employ Tucker decomposition for low-rank modeling of the higher order Hankel tensor completion.

The Tucker-based tensor completion is a non-convex optimization problem, and the existing methods usually have difficulty for selecting the step-size parameter.
In this study, we propose to use an auxiliary function-based approach, where it improves the convergence characteristics of the optimization process.
Moreover, we propose a rank increment scheme for determining the appropriate multi-linear tensor ranks.
According to our extensive experiments, the proposed method is highly suitable for tensor completion (\eg recovery of ``Lena'' from only 1\% of the randomly sampled voxels) and it outperforms state-of-the-art tensor completion methods.

\subsection{Notations}\label{sec:notation}
A vector is denoted by a bold small letter $\bm a \in \bbR{I}$.
A matrix is denoted by a bold capital letter $\bm A \in \bbR{I \times J}$.
A higher-order ($N \geq 3$) tensor is denoted by a bold calligraphic letter $\ten{A} \in \bbR{I_1 \times I_2 \times \cdots \times I_N}$.
The $i$th entry of a vector $\bm a \in \bbR{I}$ is denoted by $a_i$, and the $(i,j)$th entry of a matrix $\bm A \in \bbR{I \times J}$ is denoted by $a_{ij}$.
The $(i_1, i_2, ..., i_N)$th entry of an $N$th-order tensor $\ten{A}$ is denoted by $a_{i_1 i_2 \cdots i_N}$, where $i_n \in \{1, 2, ..., I_n\}$ and $n \in \{1, 2, ..., N\}$.
The Frobenius norm of an $N$th-order tensor is defined by $|| \ten{X} ||_F := \sqrt{\sum_{i_1, i_2, ..., i_N} x_{i_1i_2 \cdots i_N}^2}$.

A mode-$k$ unfolding (matricization) of a tensor $\ten{X}$ is denoted as $\bm X_{(k)} \in \bbR{I_k \times \Pi_{n \neq k} I_n }$.
A mode-$k$ multiplication between a tensor $\ten{X} \in \bbR{I_1 \times I_2 \times \cdots \times I_N}$ and a matrix/vector $\bm A \in \bbR{R \times I_k}$ is denoted by $\ten{Y} = \ten{X} \times_k \bm A \in \bbR{I_1 \times \cdots \times I_{k-1} \times R \times I_{k+1} \times \cdots \times I_N}$, where the entries are given by $y_{i_1 \cdots i_{k-1} r i_{k+1} \cdots i_N} = \sum_{i_k} x_{i_1 \cdots i_{k-1} i_k i_{k+1} \cdots i_N} a_{ri_k} $, and we have $\bm Y_{(k)} = \bm A \bm X_{(k)}$.

If we consider $N$ matrices $\bm U^{(n)} \in \bbR{I_n \times R_n}$ and an $N$-th order tensor $\ten{G} \in \bbR{R_1 \times R_2 \times \cdots \times R_N}$, then the multi-linear tensor product is defined as
\begin{align}
  \ten{G} \times \{\bm U\} := \ten{G} \times_1 \bm U^{(1)} \times_2 \bm U^{(2)} \cdots \times_N \bm U^{(N)}. \label{eq:multi-linear-product}
\end{align}
Moreover, a multi-linear tensor product excluding the $n$-th mode is defined as
\begin{align}
  \ten{G} \times_{-n} \{\bm U\} := \ten{G} &\times_1 \bm U^{(1)} \cdots \times_{n-1} \bm U^{(n-1)} \notag \\
                                          &\times_{n+1} \bm U^{(n+1)} \cdots  \times_N \bm U^{(N)}.
\end{align}
When we consider Tucker decomposition, $\ten{G}$ and $\bm U^{(n)}$ in Eq.~\eqref{eq:multi-linear-product} are referred to as the core tensor and factor matrices, respectively.

\section{Proposed method}\label{sec:proposed}
In this study, we assume a low-rank structure of a higher order Hankel tensor given by the MDT, which is defined in Section~\ref{sec:MDT}.  We denote this by $\mathcal{H}(\cdot)$.
The proposed method is conceptually quite simple where it comprises three steps: (1) MDT, (2) low-rank tensor approximation, and (3) inverse MDT.

Let $\ten{T} \in \bbR{I_1 \times \cdots \times I_N}$ and $\ten{Q} \in \{0,1\}^{I_1 \times \cdots \times I_N}$ be the input incomplete tensor and its mask tensor, respectively, and the first step is given by
\begin{align}
&\ten{T}_H = \mathcal{H}(\ten{T}) \in \bbR{J_1 \times \cdots \times J_M}, \\
&\ten{Q}_H = \mathcal{H}(\ten{Q}) \in \{0,1\}^{J_1 \times \cdots \times J_M},
\end{align}
where $M \geq N$.

In the second step, we obtain a low-rank approximation of $\ten{T}_H$ based on the Tucker decomposition model.
For example, we first consider the following optimization problem:
\begin{align}
  \mathop{\text{minimize}}_{\ten{G},\{ \bm U^{(m)}\}_{m=1}^M} & \ || \ten{Q}_H \circledast (\ten{T}_H - \ten{G} \times \{\bm U\} )||_F^2, \label{eq:Tucker_com}\\
  \text{s.t. } & \ \ten{G} \in \bbR{R_1 \times \cdots \times R_M}, \bm U^{(m)} \in \bbR{J_m \times R_m} (\forall m), \notag
\end{align}
where $R_m \leq J_m (\forall m)$.

Finally, the resultant tensor can be obtained by the inverse MDT of Tucker decomposition:
\begin{align}
  \widehat{\ten{X}} = \mathcal{H}^{-1}(\ten{G} \times \{\bm U\} ).
\end{align}

\subsection{MDT}\label{sec:MDT}

\subsubsection{Standard delay embedding transform}\label{sec:Hankelization}
In this section, we explain the delay embedding operation.
For simplicity, we first define a standard delay embedding transform for a vector, which can be interpreted as a time-series signal.
Let us consider a vector $\bm v = (v_1, v_2, ..., v_L)^T \in \bbR{L}$,
a standard delay embedding transform of $\bm v$ with $\tau$ is given by
\begin{align}
  {\mathcal H}_\tau(\bm v) :=
  \begin{pmatrix}
    v_1    & v_2      & \cdots & v_{L-\tau+1} \\
    v_2    & v_3      & \cdots & v_{L-\tau+2} \\
    \vdots & \vdots  & \ddots & \vdots \\
    v_\tau  & v_{\tau+1} & \cdots & v_{L} 
  \end{pmatrix} \in \bbR{\tau \times (L-\tau+1)}.
\end{align}
Thus, a standard delay embedding transform produces a duplicated matrix from a vector, where this is also referred to as ``Hankelization'' since ${\mathcal H}_\tau(\bm v)$ is a Hankel matrix.
If $\bm S \in \{0,1\}^{\tau(L-\tau+1) \times L}$ is a duplication matrix that satisfies
\begin{align}
  \text{vec}( {\mathcal H}_\tau(\bm v) ) &= \bm S \bm v,
\end{align}
then the standard delay embedding transform can be obtained by
\begin{align}
  {\mathcal H}_\tau(\bm v) &= \text{fold}_{(L,\tau)}(\bm S \bm v),
\end{align}
where $\text{fold}_{(L,\tau)} : \bbR{\tau(L-\tau+1)} \rightarrow \bbR{\tau \times (L-\tau+1)}$ is a folding operator from a vector to a matrix.

\begin{figure}[t]
\centering
\includegraphics[width=0.49\textwidth]{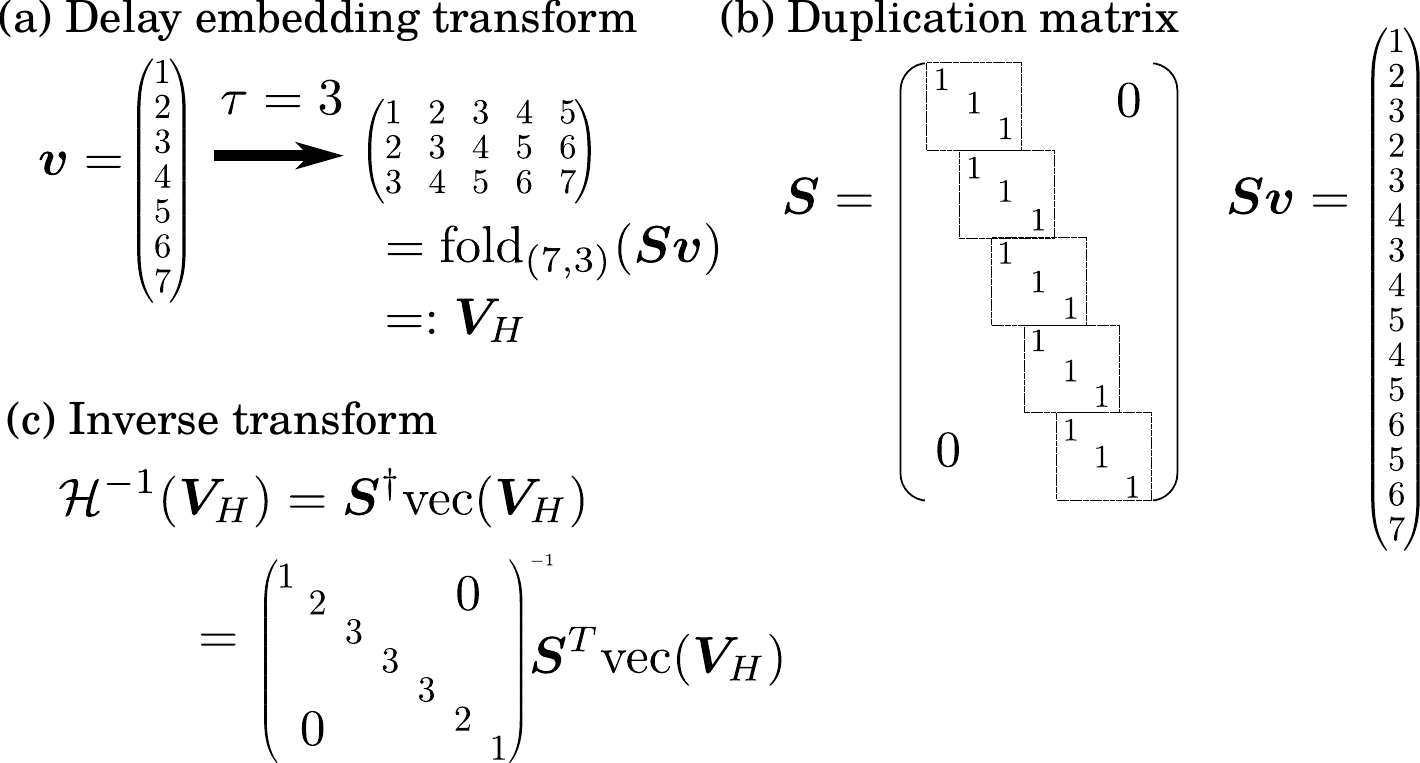}
\caption{A delay embedding transform for a vector.}\label{fig:standard_MDT}
\end{figure}

Next, we consider an inverse transform of standard delay embedding.
The forward transform can be decomposed into duplication and folding, so the inverse transform can also be decomposed into the individual corresponding inverse transforms: a vectorization operation and the Moore--Penrose pseudo-inverse $\bm S^{\dagger} := (\bm S^T \bm S)^{-1} \bm S^T$.
Thus, the inverse delay embedding transform for a Hankel matrix $\bm V_H$ can be given by
\begin{align}
  \mathcal{H}_{\tau}^{-1}(\bm V_{H}) = \bm S^{\dagger} \text{vec}(\bm V_H).
\end{align}
Figure~\ref{fig:standard_MDT} shows an example of the delay embedding transform for a vector, duplication matrix, and its inverse transform.
We can see that the duplication matrix comprises multiple identity matrices.
It should be noted that the diagonal elements of $(\bm S^T \bm S)$ comprise the numbers of duplications for individual elements, which are usually $\tau$, but low for marginal elements.

\subsubsection{Tensor extension}
\begin{figure}[t]
\centering
\includegraphics[width=0.49\textwidth]{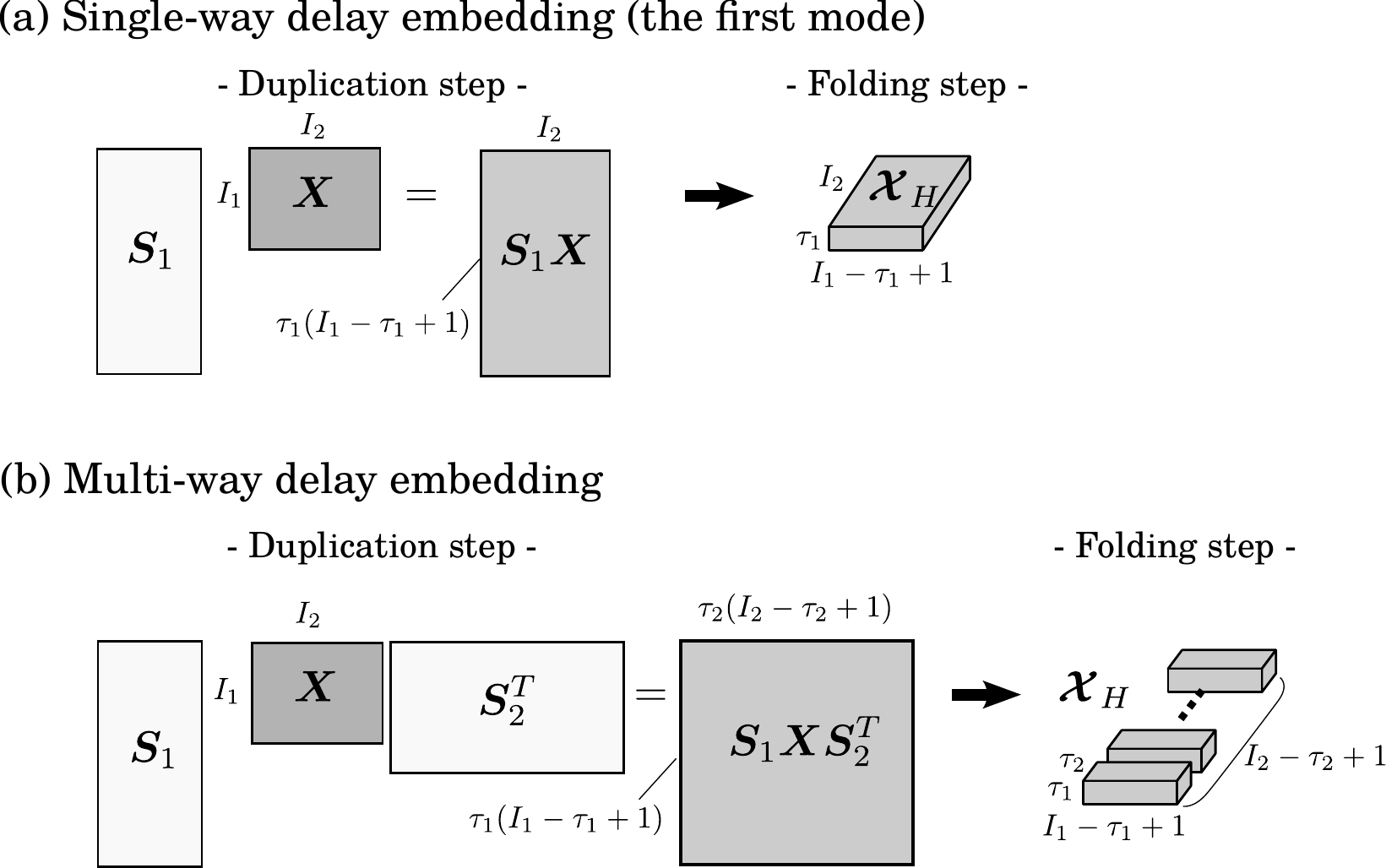}
\caption{Multi-way delay embedding transform for a matrix. Single-way and multi-way delay embedding transforms convert a matrix into third and fourth order tensors, respectively.}\label{fig:tensor_MDT}
\end{figure}

We now define the MDT for an $N$-th order tensor $\ten{X} \in \bbR{I_1 \times I_2 \times \cdots \times I_N}$.
The MDT of $\ten{X}$ with $\bm\tau \in \mathbb{N}^N$ is defined by
\begin{align}
   \mathcal{H}_{\bm \tau}(\ten{X}) = \text{fold}_{(\bm I, \bm \tau)} ( \ten{X} \times_1 \bm S_1 \cdots \times_N \bm S_N ),
\end{align}
where $\text{fold}_{(\bm I, \bm \tau)}: \bbR{\tau_1(I_1-\tau_1+1) \times \cdots \times \tau_N(I_N-\tau_N+1)}$ $\rightarrow$ $\bbR{\tau_1 \times (I_1-\tau_1+1) \times \cdots \times \tau_N \times (I_N-\tau_N+1)}$ constructs a $2N$-th order tensor from the input $N$-th order tensor.
In a similar manner to how the vector delay-embedding is a combination of linear duplication and folding operations, the MDT is also a combination of multi-linear duplication and multi-way folding operations.
Figure~\ref{fig:tensor_MDT} shows flowcharts to illustrate single-way and multi-way delay embedding for a matrix.
Finally, the inverse MDT for a Hankel tensor $\ten{X}_H$ is given by
\begin{align}
  \mathcal{H}_{\bm \tau}^{-1}(\ten{X}_H) = \text{unfold}_{(\bm I, \bm \tau)}(\ten{X}_H) \times_1 \bm S_1^\dagger \cdots \times_N \bm S_N^\dagger,
\end{align}
where $ \text{unfold}_{(\bm I, \bm \tau)} = \text{fold}_{(\bm I, \bm \tau)}^{-1}$.

\subsection{Tucker decomposition algorithms}
In this section, we explain the algorithm for solving Problem~\eqref{eq:Tucker_com}.
It should be noted that Problem~\eqref{eq:Tucker_com} is not convex, its solution is not unique, and it is not easy to obtain its global solution \cite{kolda2009tensor}.
In the case of Tucker decomposition without missing elements, it is known that the alternating least squares (ALS) \cite{de2000best} can efficiently obtain its stationary point.
In the case with missing elements, algorithms for obtaining solutions have been proposed that use the gradient descent method \cite{filipovic2015tucker} and manifold optimization \cite{kressner2014low,kasai2016low} in recent years.
Gradient descent is usually slow to converge and manifold optimization can accelerate it by correcting its update direction on the manifold.
However, a common issue with both methods is step-size parameter selection because the convergence time is sensitive to the step-size parameter.


We also propose to use an ``auxiliary function'' based approach to perform Tucker decomposition with missing elements.
The proposed algorithm is very simple but efficient because the ALS can be incorporated and it has no adjusting parameters.
First, we define the original cost function and auxiliary function by
\begin{align}
  f(\theta) &:= || \ten{Q}_H \circledast (\ten{T}_H - \ten{X}_\theta) ||_F^2, \\
  h(\theta| \theta') &:= || \ten{Q}_H \circledast (\ten{T}_H - \ten{X}_\theta) ||_F^2 \notag \\
                         &\ \ \ \ \  + || \overline{\ten{Q}}_H \circledast (\ten{X}_{\theta'} - \ten{X}_\theta) ||_F^2,
\end{align}
where $\theta = \{\ten{G}, \bm U^{(1)}, ..., \bm U^{(M)}\}$ is a set of parameters, $\overline{\ten{Q}}_H = \bm 1 - \ten{Q}_H$ represents a complement set of $\ten{Q}_H$, and $\ten{X}_\theta = \ten{G} \times \{\bm U\} $ is a Tucker decomposition.
Clearly, we have
\begin{align}
   h(\theta| \theta) = f(\theta), \text{and} \ h(\theta| \theta') \geq f(\theta) \ \ (\theta \neq \theta').
\end{align}
Let us consider the following algorithm
\begin{align}
 \theta^{k+1} = \mathop{\text{argmin}}_{\theta} h(\theta|\theta^k),
\end{align}
where the cost function is monotonically non-increasing since we have
\begin{align}
  f(\theta^{k}) = h(\theta^{k}|\theta^{k}) \geq h(\theta^{k+1}|\theta^{k}) \geq f(\theta^{k+1}).
\end{align}
It should be noted that $\theta^{k+1}$ only has to satisfy $h(\theta^{k}|\theta^{k}) \geq h(\theta^{k+1}|\theta^{k})$ to have a non-increasing property.
Furthermore, the auxiliary function can be transformed by
\begin{align}
  h(\theta| \theta^k) &= || \ten{Q}_H \circledast (\ten{T}_H - \ten{X}_\theta) ||_F^2 \notag \\
                     & \ \ \ \ \ \ \ \  + || \overline{\ten{Q}}_H \circledast (\ten{X}_{\theta^k} - \ten{X}_\theta) ||_F^2 \notag \\
                     &= || (\ten{Q}_H \circledast \ten{T}_H + \overline{\ten{Q}}_H \circledast \ten{X}_{\theta^k}) \notag \\
                     & \ \ \ \ \ \ \ \  -  (\ten{Q}_H + \overline{\ten{Q}}_H) \circledast \ten{X}_\theta||_F^2 \notag \\
                     &= || \widetilde{\ten{T}}_H^k - \ten{X}_\theta ||_F^2,
\end{align}
where $\widetilde{\ten{T}}_H^k = \ten{Q}_H \circledast \ten{T}_H +  \overline{\ten{Q}}_H \circledast \ten{X}_{\theta^k}$.
Thus, the minimization of the auxiliary function itself can be regarded as the standard Tucker decomposition without missing elements, which can be solved efficiently using the ALS.

In practice, the proposed algorithm comprises the following two steps: (1) calculate the auxiliary tensor by
\begin{align}
  \ten{Z} \leftarrow \ten{Q}_H \circledast \ten{T}_H +  \overline{\ten{Q}}_H \circledast \ten{X}_{\theta^k};
\end{align}
and (2) update $\ten{G}$ and $\{\bm U^{(m)}\}_{m=1}^{M}$ using the ALS \cite{de2000best} to optimize
\begin{align}
  \mathop{\text{minimize}}_{\ten{G}, \{\bm U^{(m)}\}_{m=1}^M} & \ || \ten{Z} - \ten{G} \times \{ \bm U \} ||_F^2, \notag \\
\text{s.t. } & \ \bm U^{(m)T} \bm U^{(m)} = \bm I_{R_m} \ \ (\forall m). \label{prob:ALS}
\end{align}
The orthogonality constraint for each $\bm U^{(m)}$ supports the uniqueness of the solution for Tucker decomposition and it does not change the reconstructed tensor from the original optimization problem.
Finally, the proposed Tucker-based tensor completion is summarized in Algorithm~\ref{alg:Tucker}.
Algorithm~\ref{alg:Tucker} updates $\bm U^{(m)}$ and $\ten{G}$ for only one cycle of the ALS, and it does not achieve a strict minimization of the auxiliary function.
However, it is guaranteed to decrease the auxiliary function because each update obtains the global minimum of the sub-optimization problem of \eqref{prob:ALS} with respect to the corresponding parameter.
Thus, Algorithm~\ref{alg:Tucker} still has the monotonic convergence property.

\begin{algorithm}[t]
\caption{Tucker-based tensor completion} \label{alg:Tucker}
\begin{algorithmic}[1]
  \STATE {\bf input:} $\ten{T} \in \bbR{J_1 \times \cdots \times J_M}$, $\ten{Q} \in \{0,1\}^{J_1 \times \cdots \times J_M}$, and $(R_1, ..., R_M)$.
  \STATE {\bf initialize:} $\ten{G} \in \bbR{R_1 \times \cdots \times R_M}$, and $\{ \bm U^{(m)} \in \bbR{J_m \times R_m} \}_{m=1}^M$, randomly.
  \REPEAT
    \STATE $\ten{X} \leftarrow \ten{G} \times \{\bm U\}$;
    \STATE $\ten{Z} \leftarrow \ten{Q} \circledast \ten{T} + \overline{\ten{Q}} \circledast\ten{X} $;
    \FOR{$m = 1, ..., M$}
      \STATE $\ten{Y} \leftarrow  \ten{Z} \times_{-m} \{\bm U^T\}$;
      \STATE $\bm U^{(m)} \leftarrow \text{$R_m$ leading singular vectors of } \bm Y_{(m)}$;
    \ENDFOR
    \STATE $\ten{G} \leftarrow \ten{Z} \times \{ \bm U^T \}$;
  \UNTIL{convergence}
  \STATE {\bf output:} $\ten{G}, \bm U^{(1)}, ..., \bm U^{(M)}$;
\end{algorithmic}
\end{algorithm}

\subsection{Tucker decomposition with rank increment}
\begin{figure}[t]
\centering
  \includegraphics[width=0.49\textwidth]{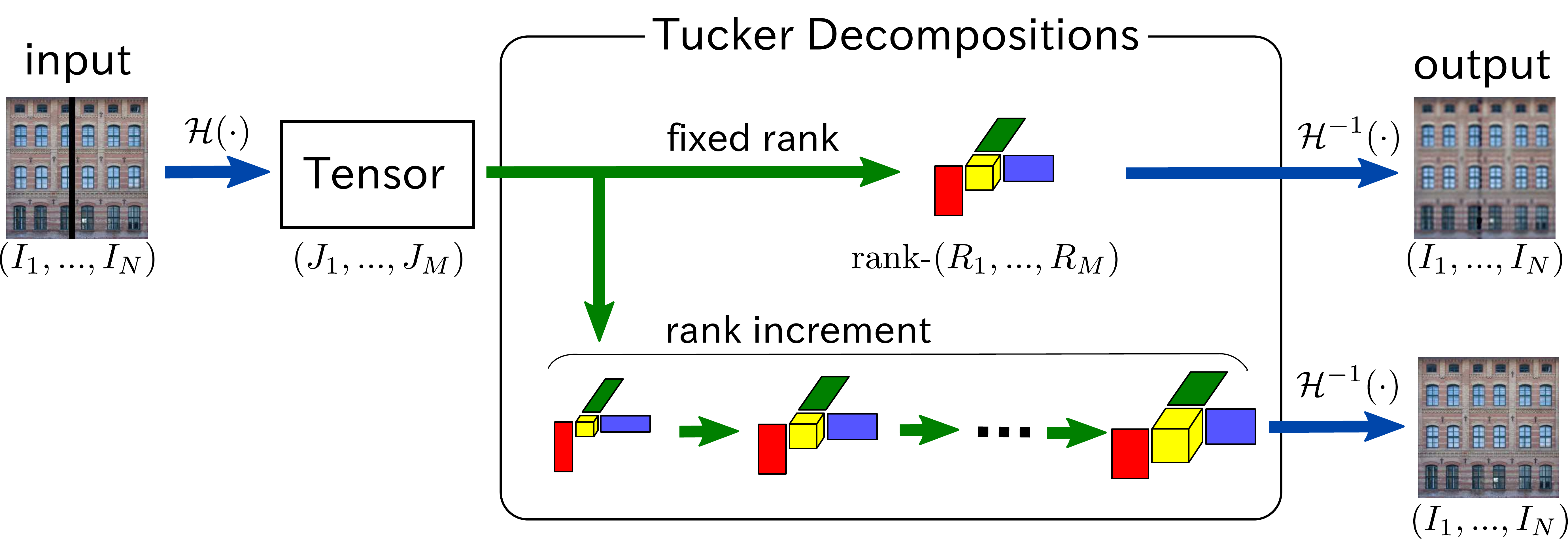}
  \caption{Conceptual illustrations of the proposed methods.}\label{fig:concept}
\end{figure}

A difficult and important issue with the Tucker-based tensor completion method is determining an appropriate rank setting $(R_1, ..., R_{M})$.
If we aim to obtain the lowest rank setting for sufficient approximation, the rank estimation problem can be considered as
\begin{align}
  \mathop{\text{minimize}}_{(R_1, ..., R_{M})} & \ \sum_m R_m, \notag \\ 
  \text{s.t. }& \ || \ten{Q}_H \circledast (\ten{T}_H - \ten{X}) ||_F^2 \leq \epsilon, \label{eq:rank_est}\\
              & \ \text{rank}(\ten{X}) = (R_1, ..., R_{M}) \notag,
\end{align}
where $\epsilon$ is a noise threshold parameter.
However, we do not know the existence of the unique solution $(R_1^*, ..., R_M^*)$ for Problem~\eqref{eq:rank_est} and it will be dependent on $\epsilon$.
Furthermore, even if the best rank setting is unique, the resultant low-rank tensor $\ten{X}^*$ is not unique.
To address this issue, we propose the use of a very important strategy called the ``rank increment'' method.
Figure~\ref{fig:concept} provides a flowchart to illustrate the concept of Tucker decomposition with/without rank increment.
The rank increment strategy has been discussed in several studies of matrix and tensor completion \cite{mishra2013low,dolgov2013alternating,tan2014riemannian,uschmajew2015greedy,yokota2015a,yokota2016smooth}, but the present study is its first application to Tucker-based completion to the best of our knowledge.
The main reason for using the rank increment method is the non-uniqueness of the solution for the tensor $\ten{X}$.
Thus, the resultant tensor depends on its initialization.
The main feature of the rank increment method is that the tensor should be initialized by a lower rank approximation than its target rank.
Based on this strategy, the proposed algorithm can be described as follows. 
\begin{itemize}\setlength{\leftskip}{0.1cm}\setlength\itemsep{1mm}
  \item{Step 1:} Set initial $R_m=1$ for all $m$.
  \item{Step 2:} Obtain $\ten{G}$ and $\{\bm U ^{(m)}\}_{m=1}^M$ with $(R_1, ..., R_{M})$ using Algorithm~\ref{alg:Tucker} and obtain $\ten{X} = \ten{G} \times \{\bm U\}$.
  \item{Step 3:} Check the noise condition $|| \ten{Q}_H \circledast (\ten{T}_H - \ten{X}) ||_F^2 \leq \epsilon$, where the algorithm is terminated if it is satisfied; otherwise, go to the next step.
  \item{Step 4:} Choose the incremental mode $m'$ and increment $R_{m'}$, and then go back to step 2.
\end{itemize}
The problem is how to choose $m'$ and how to increase the rank $R_{m'}$.
We propose choosing $m'$ using the ``$m$-th mode residual'' of the cost function, which is defined as a residual on the multi-linear subspace spanned by all the factor matrices excluding the $m$-th mode factor.
This is mathematically formulated as:
\begin{align}
  m' = \mathop{\text{argmax}}_m || (\ten{Q}_H \circledast (\ten{T}_H - \ten{X})) \times_{-m} \{ \bm U^T \}  ||_F^2.
\end{align}
We can interpret this as meaning that the selected $m'$-th mode has a high expectation of cost reduction when $R_{m'}$ increases while the other-mode ranks remain fixed.

For the rank increment process, we consider the rank sequences for individual modes.
For example, the rank sequence for the $m$-th mode is set as $ \bm L_m = [1, 2, 4, 8, ..., J_m]$ because the contribution rates of the singular vectors usually decrease exponentially.
Thus, a small rank increment is important for the phase of a low-rank approximation, whereas a small rank increment is not effective for the phase of a relatively high-rank approximation.
Large rank steps for the high-rank phase help to accelerate the algorithm, but they should be selected carefully because excessively large rank steps may lead to problems with non-unique solutions.
The proposed method for Tucker-based tensor completion with rank increment is summarized in Algorithm~\ref{alg:Tucker_with_rank}.

\begin{algorithm}[t]
\caption{Tucker-based tensor completion with rank increment} \label{alg:Tucker_with_rank}
\begin{algorithmic}[1]
  \STATE {\bf input:} $\ten{T} \in \bbR{J_1 \times \cdots \times J_M}$, $\ten{Q} \in \{0,1\}^{J_1 \times \cdots \times J_M}$, $\{\bm L_1, ..., \bm L_M\}$, $\epsilon$, tol.
  \STATE {\bf initialize:} $k_m \leftarrow 1, R_m \leftarrow \bm L_m(k_m)$ \  $(\forall m)$, $\ten{G} \in \bbR{R_1 \times \cdots \times R_M}$, and $\{ \bm U^{(m)} \in \bbR{J_m \times R_m} \}_{m=1}^M$;
  \STATE $\ten{X} \leftarrow \ten{G} \times \{\bm U\}$;
  \STATE $f_1 \leftarrow || \ten{Q} \circledast (\ten{T}-\ten{X}) ||_F^2$;
  \REPEAT
    \STATE Do lines 5-10 in Algorithm~\ref{alg:Tucker};
    \STATE $\ten{X} \leftarrow \ten{G} \times \{\bm U\}$;
    \STATE $f_2 \leftarrow || \ten{Q} \circledast (\ten{T}-\ten{X}) ||_F^2$;
    \IF{ $|f_2 - f_1| \leq$ tol}
      \STATE $\widetilde{\ten{X}} \leftarrow \ten{Q} \circledast (\ten{T} - \ten{X})$;
      \STATE $m' \leftarrow \mathop{\text{argmax}}_m || \widetilde{\ten{X}} \times_{-m} \{ \bm U^T \}  ||_F^2$;
      \STATE $k_{m'} \leftarrow k_{m'} + 1$, and $R_{m'} \leftarrow \bm L_{m'}(k_{m'})$;
    \ELSE
      \STATE $f_1 \leftarrow f_2$;
    \ENDIF
  \UNTIL{$f_2 \leq \epsilon$}
  \STATE {\bf output:} $\ten{G}, \bm U^{(1)}, ..., \bm U^{(M)}$;
\end{algorithmic}
\end{algorithm}

\section{Experiments}\label{sec:experiments}


\subsection{Verification of the proposed method using a typical color image}
\begin{figure}[t]
  \centering
  \includegraphics[width=0.49\textwidth]{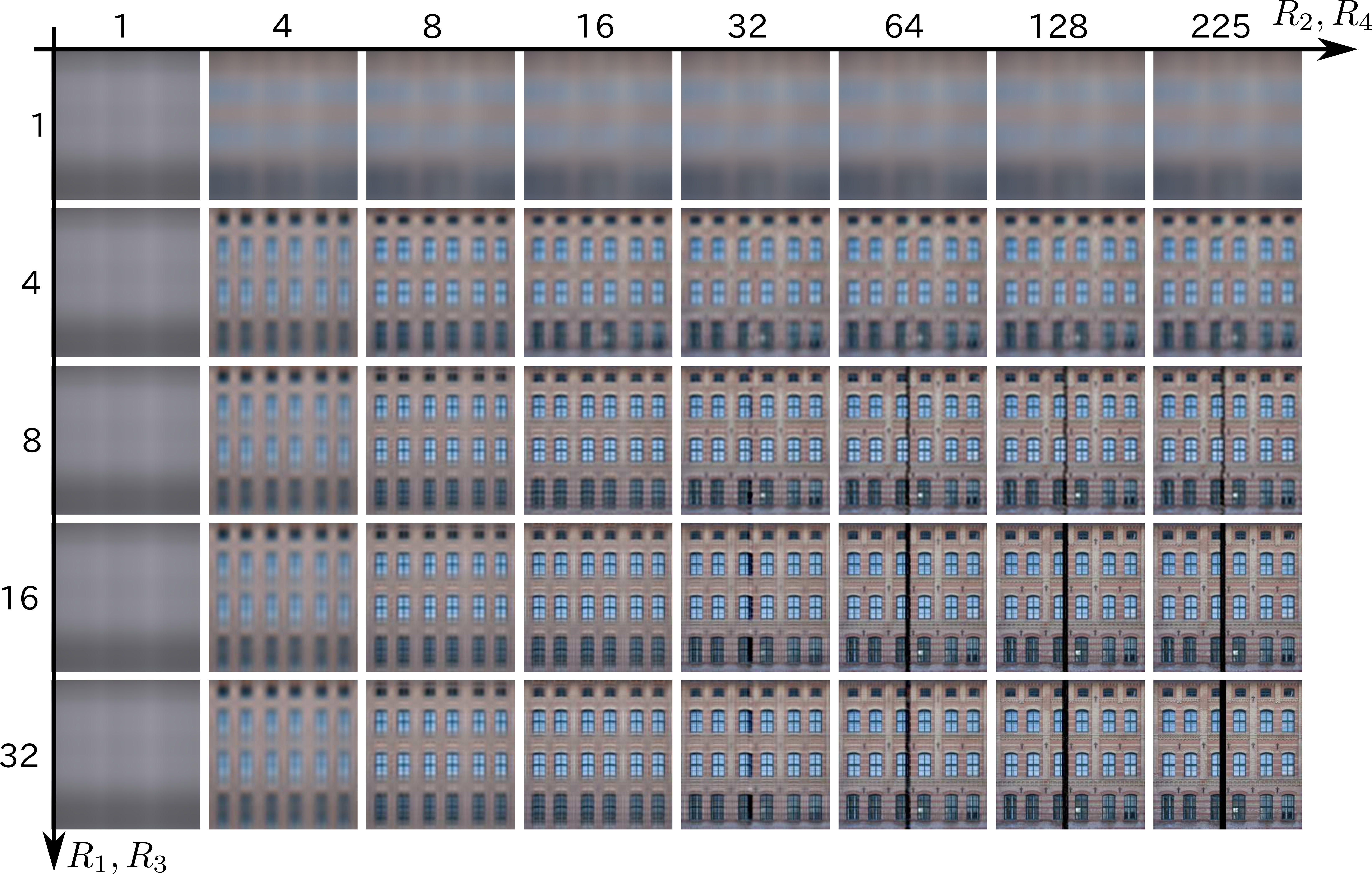}
  \caption{Results obtained by fixed rank MDT Tucker decomposition with $\bm\tau=(32,32,1)$ for various rank settings.}\label{fig:result_fixed}
  \vspace{5mm}
  \centering
  \includegraphics[width=0.49\textwidth]{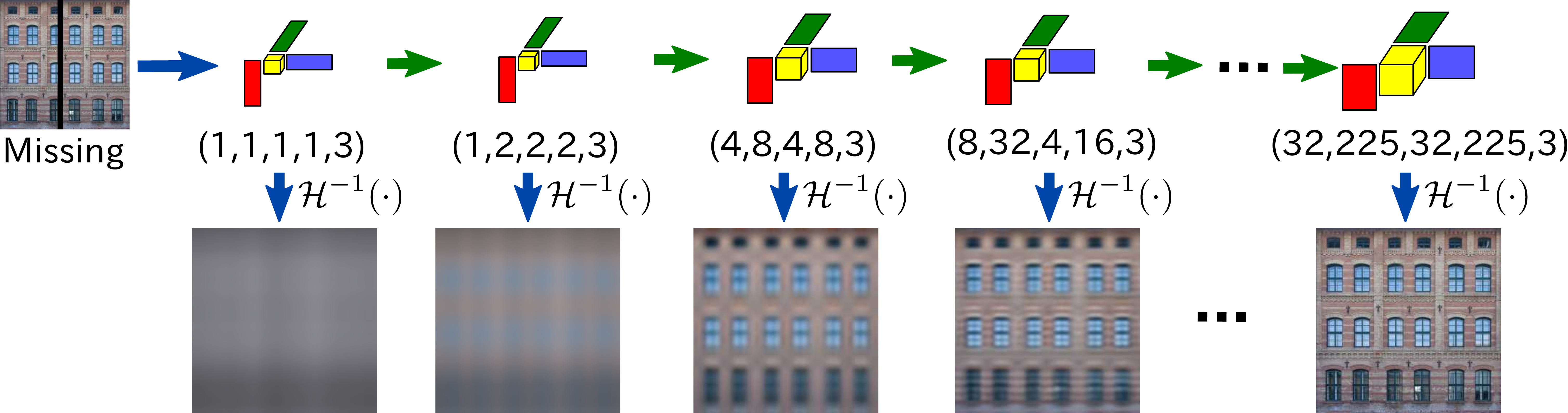}
  \caption{Results obtained by rank increment MDT Tucker decomposition with $\bm\tau=(32,32,1)$.}\label{fig:result_incR}
\end{figure}

First, in our experiments, we tried to fill the missing slices in a typical color image by using MDT and fixed rank Tucker decomposition.
The input image is depicted in Figure~\ref{fig:concept}.
We set $\bm \tau=(32,32,1)$ and a $(256,256,3)$ color image was converted into a $(32,225,32,225,1,3)$ tensor.
The fifth mode can be ignored so this Hankel tensor was regarded as a fifth-order tensor with a size of $(32,225,32,225,3)$.
Figure~\ref{fig:result_fixed} shows the images obtained with various settings for the rank parameter.
Clearly, low-rank Tucker decomposition with the Hankel tensor successfully filled the missing area.
However, an important issue is how to treat the difference between the meanings of $(R_1,R_3)$ and $(R_2,R_4)$.
The fundamental difference between $(R_1,R_3)$ and $(R_2,R_4)$ is due to the window sizes of $32$ and $225$.
Thus, it should be noted that a lower $(R_1,R_3)$ may contribute to the representation of the local structure (e.g., smoothness), whereas a lower $(R_2,R_4)$ may contribute to the representation of the global structure (e.g., recursive textures) of the image.
In Figure~\ref{fig:result_fixed}, when we compare two flows from the bottom right to the top right, and from the bottom right to the bottom left, the low-rankness of $(R_2,R_4)$ is clearly more important than that of $(R_1,R_3)$ for recovering the missing area.


Next, we tried to fill missing color images using MDT and Tucker decomposition with the rank increment method.
The rank sequences for the proposed method were set as $\bm L_1 = \bm L_3 = (1,2,4,8,16,24,32)$, $\bm L_2=\bm L_4=(1,2,4,8,16,32,64,96,128,160,192,225)$, and $\bm L_5=3$.
Figure~\ref{fig:result_incR} shows the main flow for the processed images using the proposed method.
Using the rank increment method, we first obtained a Tucker decomposition with a very low-rank setting (e.g., rank-one tensor) and a higher-rank decomposition was then obtained by using the previous lower-rank decomposition for its initialization.
We repeatedly obtained a higher-rank decomposition until the residual was sufficiently small.
Thus, the rank increment method automatically selected an appropriate rank setting.

\begin{figure*}[t]
  \centering
  \includegraphics[width=0.99\textwidth]{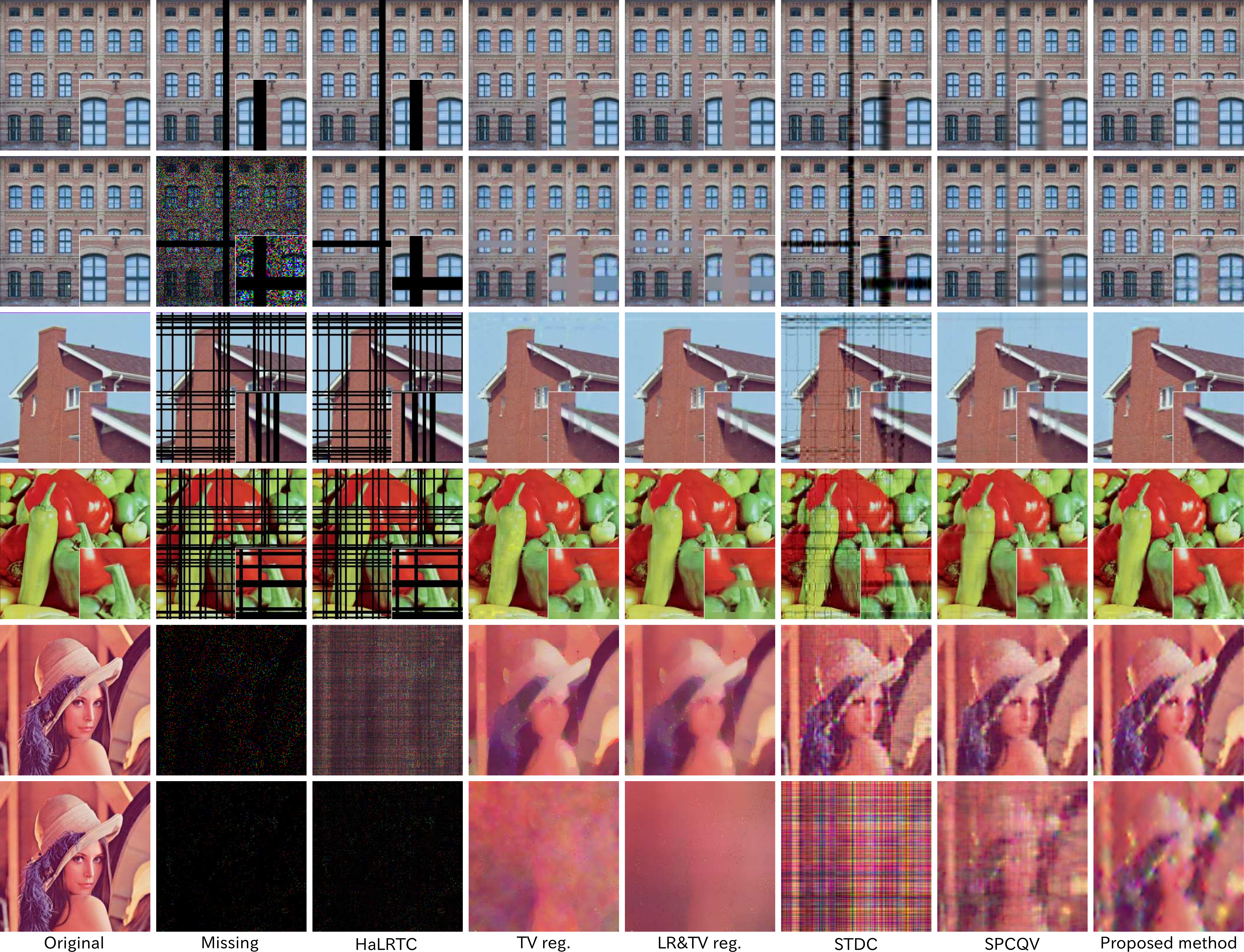}
  \caption{Color images completed with various methods.  Six missing color images were artificially generated comprising: slice missing case ``facade 1,'' slice+voxel missing case ``facade 2,'' random slice missing cases ``house' and ``peppers,'' and random voxel missing cases ``Lena (95\%)' and ``Lena (99\%)'. }\label{fig:facade_comp}
\end{figure*}

\begin{table*}[t]
  \centering
  \caption{ Comparisons of the peak signal-to-noise ratio (PSNR) and structural similarity (SSIM) after color image completion.}\label{tab:comparison}
  \begin{tabular}{l | l l l l l l} \hline
  (PSNR,SSIM)& HaLRTC        & TV reg.       & LR\&TV reg. & STDC & SPCQV & Proposed \\ \hline
  facade 1  & (19.31,0.937) &(30.08,0.964) &(30.18,0.964) &(24.03,0.958) &(30.04,0.972) &({\bf 36.52},{\bf 0.988}) \\
  facade 2  &(16.03,0.852) &(24.28,0.838) &(24.47,0.842) &(17.89,0.826) &(25.65,{\bf 0.916}) &({\bf 26.96},{\bf 0.916}) \\
  house     &(8.81,0.271) &(26.75,0.909) &(26.70,0.913) &(20.65,0.624) &(27.04,{\bf 0.909}) &({\bf 27.50},0.908) \\
  peppers   &(9.92,0.288) &(25.84,0.877) &(25.81,0.885) &(21.61,0.662) &(26.59,0.888) &({\bf 27.62},{\bf 0.898}) \\ 
  Lena (5\%)&(9.63,0.145) &(20.86,0.626) &(20.89,0.621) &(22.07,0.596) &(23.52,0.700) &({\bf 23.57},{\bf 0.738}) \\ 
  Lena (1\%)&(5.32,0.021) &(15.47,0.440) &(15.49,0.443) &(12.75,0.089) &(18.49,0.508) &({\bf 19.68},{\bf 0.565}) \\
\hline
  \end{tabular}
\end{table*}

\begin{figure}[t]
  \centering
  \includegraphics[width=0.49\textwidth]{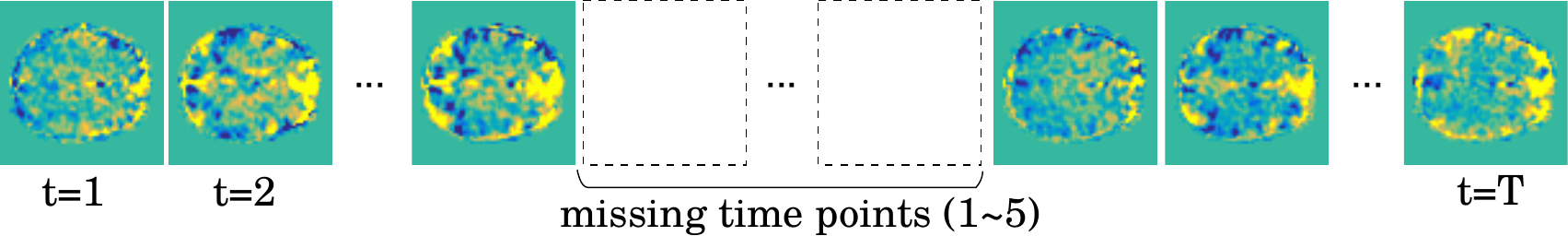}
  \caption{Missing time frames in functional magnetic resonance images.}\label{fig:fMRI_data}
\centering
\vspace{5mm}
 \includegraphics[width=0.49\textwidth]{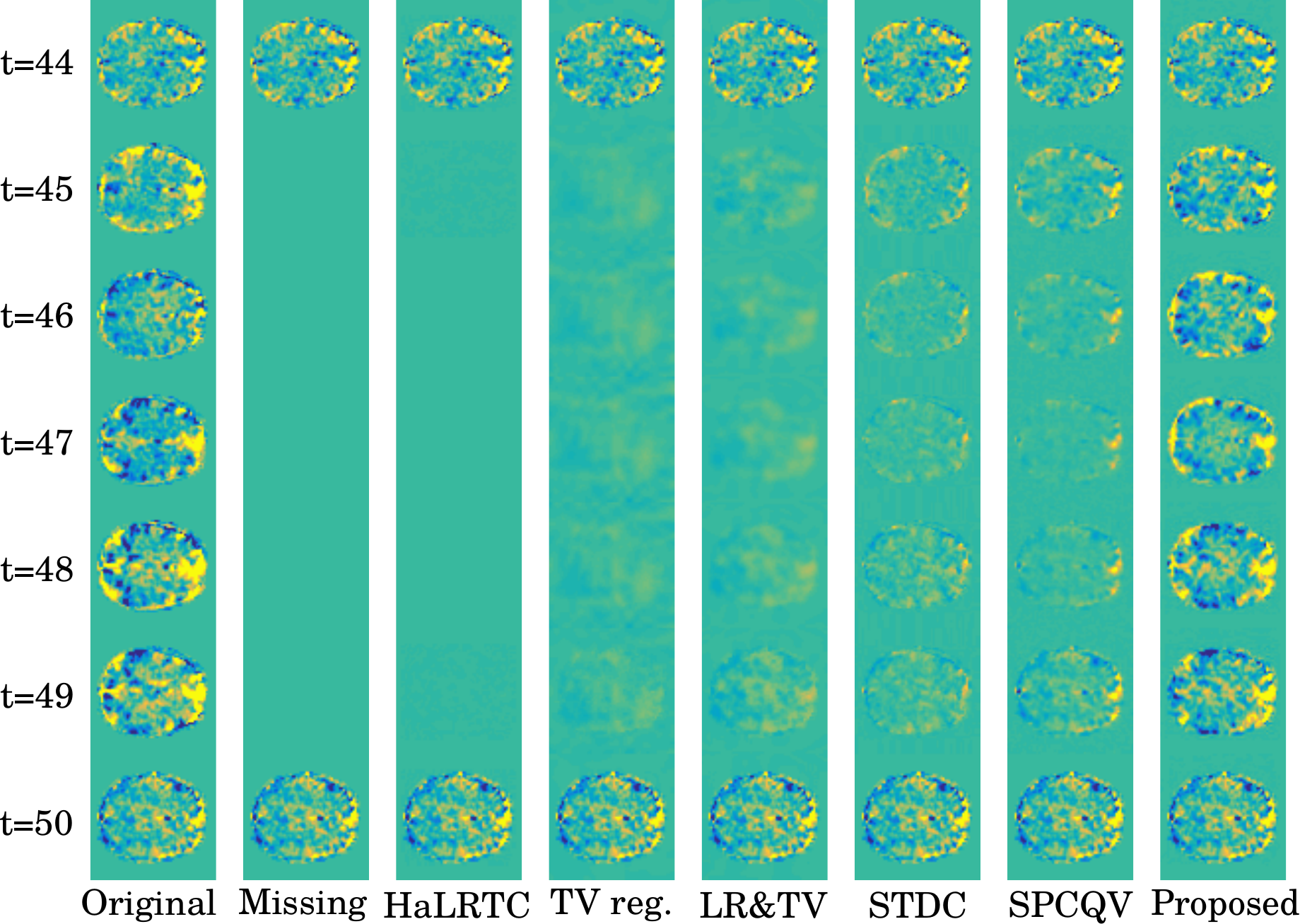}
\caption{Results recovered for functional magnetic resonance image slices.}\label{fig:result_fMRI_slices}
\end{figure}

\subsection{Comparison using color images}
We compared the performance of the proposed method with those of state-of-the-art tensor completion algorithms: HaLRTC (nuclear-norm regularization) \cite{liu2013tensor}, TV regularization \cite{zhu2008efficient}, nuclear-norm and TV regularization (LR\&TV) \cite{yokota2017simultaneous}, STDC (constrained Tucker decomposition) \cite{chen2014simultaneous}, and SPCQV (constrained PARAFAC tensor decomposition) \cite{yokota2016smooth}.
We prepared six missing images for this experiment.
The first image had 11 continuous missing slices along the vertical axis.
Several horizontal and vertical slices and many voxels were missing from the second image.
Random vertical and horizontal slices were missing from the third and fourth images.
In addition, 95\% and 99\% of the random voxels were missing from the fifth and sixth images, respectively.

Figure~\ref{fig:facade_comp} shows the experimental results obtained after the completion of various incomplete images.
Magnified regions are depicted at the bottom right in the first to fourth images.
HaLRTC recovered random missing voxels for the second image, but it failed to recover the missing slices for all of the images.
The TV regularization and LR\&TV regularization methods filled the missing areas, but the recovered areas were unnaturally flat.
STDC failed to recover the missing slices and SPCQV retained ``shadows'' of the missing slices.
By contrast, the proposed method recovered most of the missing slices without shadows.
For the image with 95\% missing voxels, the proposed method and SPCQV obtained similar results.
However, for the image with 99\% missing voxels, the result obtained by the proposed method was much better than that by SPCQV.
Table~\ref{tab:comparison} shows the peak signal-to-noise ratio (PSNR) and structural similarity (SSIM) \cite{wang2004image} for these comparisons, where the best PSNR and SSIM values are emphasized in bold font.
According to this quantitative evaluation, the proposed method performed better than the state-of-the-art methods in terms of the PSNR, and the results in terms of the SSIM were very competitive with SPCQV for some of the images.


\begin{table*}[t]
\caption{ SNR and mean SSIM after the completion of fMRI slices.}\label{tab:fMRI_comp}
\centering
\begin{tabular}{ l | l l l l l l} \hline
 \# of missing slices & HaLRTC        & TV reg.       & LR\&TV reg. & STDC & SPCQV & Proposed \\ \hline
1  & (20.15,0.991) &(23.13,0.992) &(23.16,0.993) &(23.30,0.993) &({\bf 24.20},{\bf 0.994}) &(23.25,{\bf 0.994}) \\
2  & (14.71,0.982) &(16.20,0.982) &(16.29,0.984) &(17.09,0.985) &(17.80,0.985) &({\bf 18.45},{\bf 0.987}) \\
3  & (14.02,0.973) &(15.20,0.972) &(15.17,0.974) &(15.99,0.977) &({\bf 16.58},0.976) &(14.54,{\bf 0.978}) \\
4  & (12.17,0.964) &(12.45,0.962) &(12.46,0.964) &(12.43,0.966) &(12.98,0.966) &({\bf 14.08},{\bf 0.970}) \\
5  & (11.31,0.955) &(11.41,0.952) &(11.59,0.956) &(11.59,0.957) &(12.15,0.957) &({\bf 13.85},{\bf 0.964}) \\ \hline
\end{tabular}
\end{table*}

\subsection{Comparison using fMRI images }
In the next experiment, we tried to recover continuous missing time frames in fMRI images.
Figure~\ref{fig:fMRI_data} shows the image prepared with missing data.
There were 94 time frames of the fMRI slices and one to five continuous time frames were removed.
The size of the tensor was $(64 \times 64 \times 94)$.
We applied the proposed method with $\bm\tau=(2,2,32)$ for the given tensor.
The rank sequences were set as $\bm L_1 = \bm L_3 = (1,2)$, $\bm L_2=\bm L_4=\bm L_6=(1,2,4,8,16,32,63)$, and $\bm L_5=(1,2,4,8,16,32)$.


Figure~\ref{fig:result_fMRI_slices} shows the results of this experiments.
Similar to the color image experiments, the ordinary low-rank model could not recover the missing slices.
TV regularization obtained flat results and LR\&TV regularization produced similar results.
STDC and SPCQV obtained some improvements compared with the convex methods, but they were still unclear. 
By contrast, the results obtained by the proposed method were very clear, although their accuracy was not high.
Table~\ref{tab:fMRI_comp} shows the signal-to-noise ratio (SNR) and mean SSIM results obtained for all of the completion methods, which demonstrates that the proposed method performed better than the state-of-the-art methods in terms of the mean SSIM measure and it was also very competitive in terms of the SNR.

\section{Discussion}\label{sec:dis}
\subsection{Novelty and contributions}
\subsubsection{Low-rank model in embedded space}
The idea of tensor completion using MDT and its inverse transform is novel.
Most of the existing methods for tensor completion are based on structural assumptions in the signal space, such as nuclear-norm regularization and TV regularization.
By contrast, our method considers a structural assumption in the embedded space, which can be regarded as a novel strategy for the tensor completion problem.
Furthermore, we employ delay-embedding in this approach and it is extended it in a multi-way manner for tensors.

\subsubsection{An auxiliary function-based approach for Tucker decomposition of a tensor with missing elements}
The auxiliary function-based algorithm for Tucker decomposition is efficiently employed.
The existing algorithms used for the Tucker decomposition of a tensor with missing elements \cite{filipovic2015tucker,kressner2014low,kasai2016low} are based on gradient methods.
However the convergence speed of the gradient method is quite sensitive to the step-size parameter.
By contrast, the proposed algorithm does not have any hyper-parameters and its monotonic convergence is guaranteed by the auxiliary function theory.

\subsubsection{Model selection and uniqueness improvement by the rank increment method}
The rank increment method for the Tucker decomposition of incomplete tensors is firstly applied in the best of our knowledge.
Several methods for estimating multi-linear tensor ranks have been studied only for complete tensors \cite{timmerman2000three,yokota2017robust}.
Methods for estimating the ranks of matrix and PARAFAC decompositions have been proposed for incomplete data \cite{tan2014riemannian,uschmajew2015greedy,yokota2015a,yokota2016smooth}, but these methods cannot be applied to our problem.
Thus, we proposed a new method for estimating multi-linear ranks for an incomplete tensor, where it can also handle the issue of non-unique solutions.


\subsection{Computational bottleneck}
The proposed method has an issue with data volume expansion due to MDT.
An $N$-th order tensor is converted into a $2N$-th order tensor by MDT and its data size increases roughly $\prod_{n=1}^N \tau_n$-fold.
This issue makes it difficult to apply the proposed method to large-scale tensors and this problem will be addressed in future research.

\section{Conclusions}\label{sec:con}
In this study, we proposed a novel method and algorithm for tensor completion problems that include missing slices.
The recovery of missing slices is recognized as a difficult problem that ordinary tensor completion methods usually fail to solve.
To address this problem, we introduced the concept of ``delay embedding'' from the study of dynamical systems and extended it for our problem.
We showed that missing slices can be recovered by considering low-rank models in embedded spaces and that MDT is a good choice for this purpose.

At present, the proposed method is very basic but it has many potential extensions such as using different embedding transformations and constrained tensor decompositions (e.g., non-negative, sparse, and smooth).
The MATLAB code will be available via our website\footnote{\url{https://sites.google.com/site/yokotatsuya/home/software}}.





\clearpage
\clearpage
\newpage

{\small

}

\clearpage
\clearpage
\newpage

\section*{Supplemental Figures: Results for Various Color-image Completion}

In this supplemental document, we show the results of color-image completion by using various methods with explanations of the technical overview of the state-of-the-art methods.
Table~\ref{tab:opt_problems} summarizes the optimization concepts of the state-of-the-art tensor completion methods: high accuracy low-rank tensor completion (HaLRTC) \cite{liu2013tensor}, total variation regularization (TV reg.) \cite{zhu2008efficient,yokota2017simultaneous}, low-rank and TV regularization (LRTV reg.) \cite{yokota2017simultaneous}, simultaneous tensor decomposition and completion \cite{chen2014simultaneous}, smooth PARAFAC tensor completion with quadratic variation (SPCQV) \cite{yokota2016smooth}, and the proposed method.

In HaLRTC, we minimize the tensor nuclear norm (tNN), which is defined as sum of matrix nuclear norm (NN) for all $n$-th matricizations of $\ten{X}$, under the constraint of data consistency $\ten{P}_\Omega(\ten{T})= \ten{P}_{\Omega}(\ten{X})$.
In TV reg., we minimize the tensor total variation (tTV), which is an extension of matrix total variation (TV), under the constraint of data consistency.
LRTV simultaneously minimizes tNN and tTV under the constraint of data consistency.
Above three methods are formulated as convex optimization problems, and can be solved by primal-dual splitting (PDS) algorithm \cite{condat2013primal}.
Since LRTV method proposed in \cite{yokota2017simultaneous} is a generalization of both HaLRTC and TV reg. methods and PDS algorithm is employed for optimization, we tried above three methods via PDS algorithm with appropriate hyper-parameter settings in LRTV formulation.
In STDC, a Tucker decomposition problem with regularizations for a core tensor and factor matrices is considered.
Factor matrices in STDC are imposed to small nuclear-norm to minimize the tensor rank, and factor prior imposes a kind of similarity of factor vectors each other.
The optimization problem of STDC is non-convex, and it is solved by using augmented Lagrangian method.
According to \cite{chen2014simultaneous}, the convergence of STDC algorithm is not theoretically guaranteed.
In SPCQV, a PARAFAC decomposition problem with regularizations for factor vectors is considered.
The PARAFAC decomposition consists of multiple rank-1 tensors, and each rank-1 tensor is constructed by $N$ factor vectors.
In SPCQV method, individual factor vectors are imposed to be smoothly variated, and number of rank-1 tensors is minimized.
The optimization problem of SPCQV is also non-convex, and it is solved by using hierarchical alternating least squares (HALS) algorithm.
Unlike the STDC, SPCQV has a monotonic non-increasing property of cost function and the convergence to stationary point is guaranteed.
In the proposed method, a low-rank Tucker decomposition problem in embedded space is simply considered.
In general, the optimization problem is non-convex, however, monotonic convergence property is guaranteed based on auxiliary function based optimization algorithm.

\begin{figure*}[t]
\centering
\includegraphics[width=0.99\textwidth]{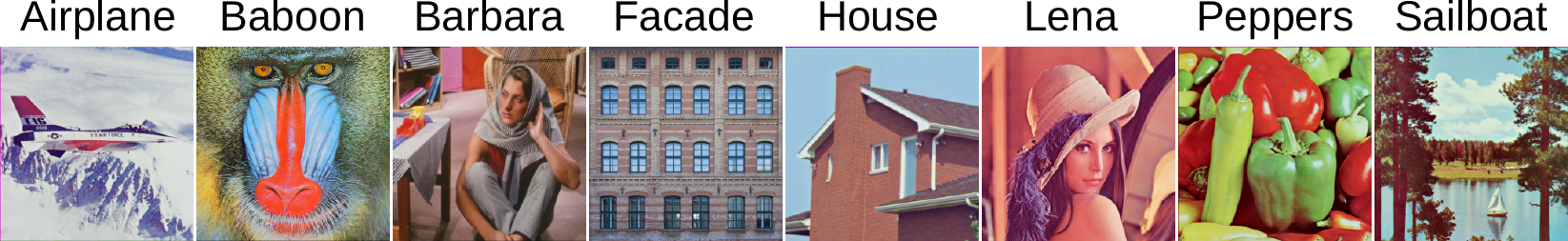}
\caption{Eight benchmark images}\label{fig:test_images}
\end{figure*}

\begin{figure*}[t]
\centering
\includegraphics[width=0.99\textwidth]{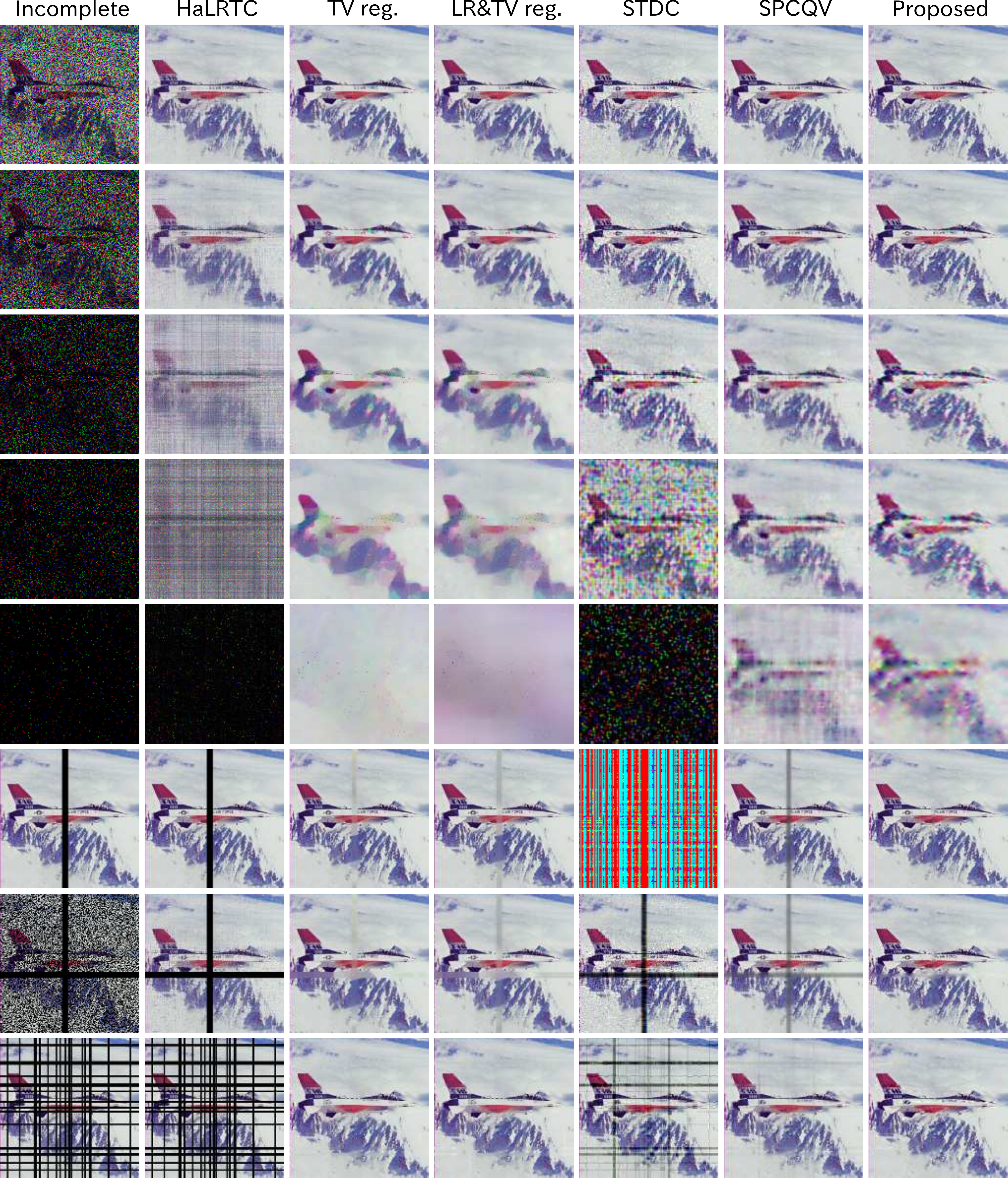}
\caption{Results in `Airplain'}\label{fig:airplain}
\end{figure*}

\begin{figure*}[t]
\centering
\includegraphics[width=0.99\textwidth]{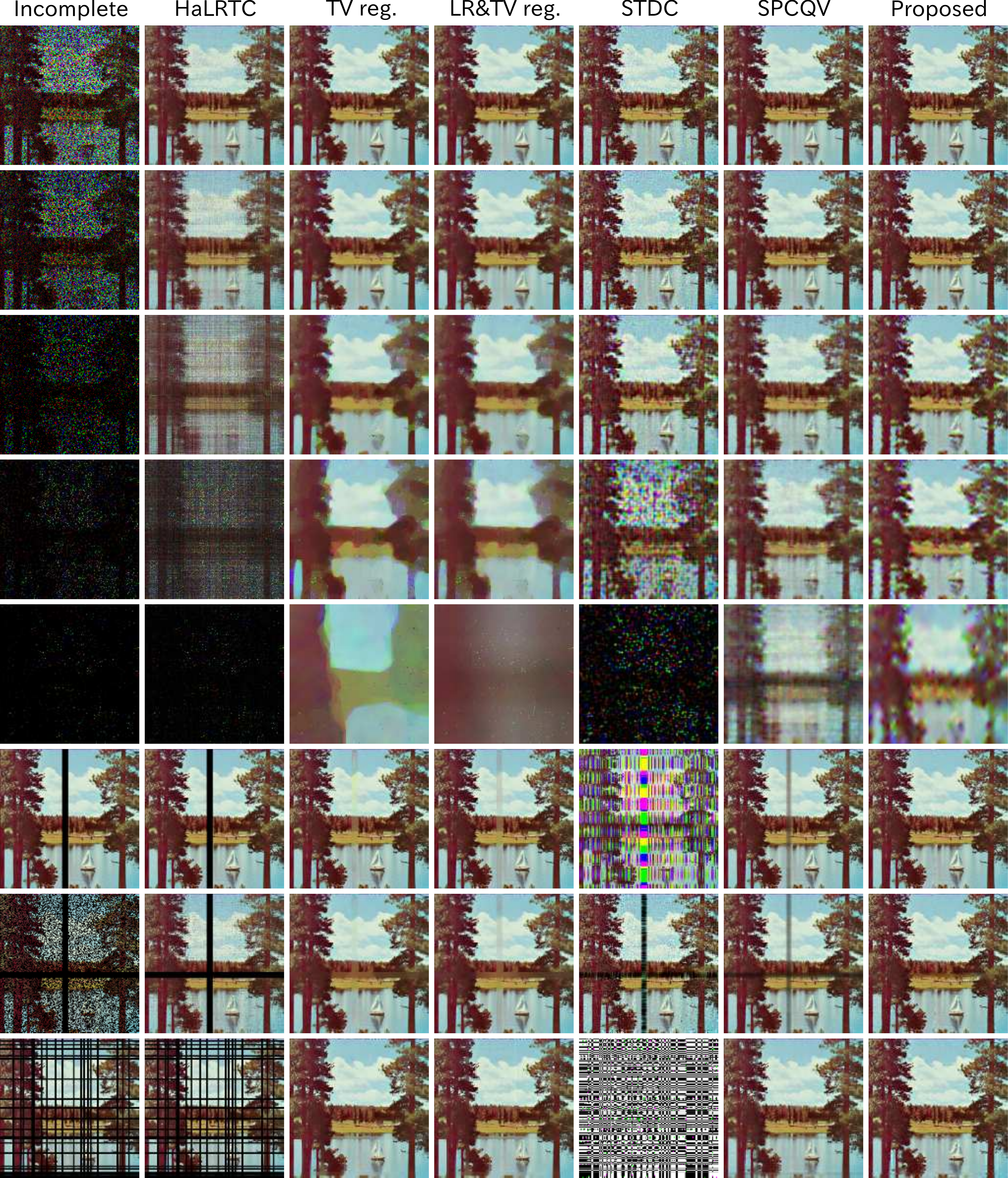}
\caption{Results in `Sailboat'}\label{fig:sailboat}
\end{figure*}

Figure~\ref{fig:test_images} shows eight benchmark images used in this experiments.
For each image, we generate eight types of incomplete images: (a) 50\%, (b) 70\%, (c) 90\%, (d) 95\%, and (e) 99\% random voxel missing, (f) 11 continuous vertical slices missing, (g) cross shape occlusion with 50\% random voxel missing, and (h) random vertical/horizontal slices missing.
Thus, totally, 64 incomplete images were generated.
For each incomplete image, we applied six tensor completion methods: HaLRTC, TV reg., LRTV reg., STDC, SPCQV, and the proposed method.
Hyper parameters for all methods were tuned from several candidates, and employed the best settings.
Figures~\ref{fig:airplain}-\ref{fig:sailboat} show the results of this experiments.
First three convex methods recovered them for only low missing ratio cases.
STDC outperformed the three convex methods, however, it failed to recover 95\% and 99\% missing cases and slice missing cases, and sometimes the algorithm did not converge.
SPCQV successfully recovered random missing cases, however, it failed to recover slice missing cases.
The proposed method obtained more clear results for random missing cases compared with SPCQV, and successfully recovered slice missing cases.

\begin{table*}[h]
\centering
\caption{Optimization concepts}\label{tab:opt_problems}
{\footnotesize
\begin{tabular}{ l l l } \hline
 name & minimization & constraints \\ \hline
 HaLRTC \cite{liu2013tensor}    & tensor nuclear norm (tNN): $||\ten{X}||_{LR}$ & $\ten{P}_{\Omega}(\ten{T}) = \ten{P}_{\Omega}(\ten{X})$ \\
 TV reg. \cite{zhu2008efficient,yokota2017simultaneous}   & tensor total variation (tTV): $||\ten{X}||_{TV}$ & $\ten{P}_{\Omega}(\ten{T}) = \ten{P}_{\Omega}(\ten{X}) $ \\
 LRTV reg. \cite{yokota2017simultaneous} & (tNN) + (tTV) & $\ten{P}_{\Omega}(\ten{T}) = \ten{P}_{\Omega}(\ten{X}) $  \\
 STDC \cite{chen2014simultaneous}      & (NN of factor matrices) + (l2-norm of core tensor) + (factor prior) & Tucker decomposition + $\ten{P}_{\Omega}(\ten{T}) = \ten{P}_{\Omega}(\ten{X}) $ \\
 SPCQV \cite{yokota2016smooth}   & (number of rank-1 tensors) + (quadratic variation of factor vectors) & PARAFAC decomposition + $\ten{P}_{\Omega}(\ten{T}) = \ten{P}_{\Omega}(\ten{X}) $\\
 Proposed          & (sum of multi-linear tensor ranks of $\ten{X}_H$) & $\ten{P}_{\Omega}(\ten{T}) = \ten{P}_{\Omega}(\mathcal{H}^{-1}(\ten{X}_H))$ \\ \hline
\end{tabular}
}
\end{table*}

\begin{figure*}[t]
\centering
\includegraphics[width=0.99\textwidth]{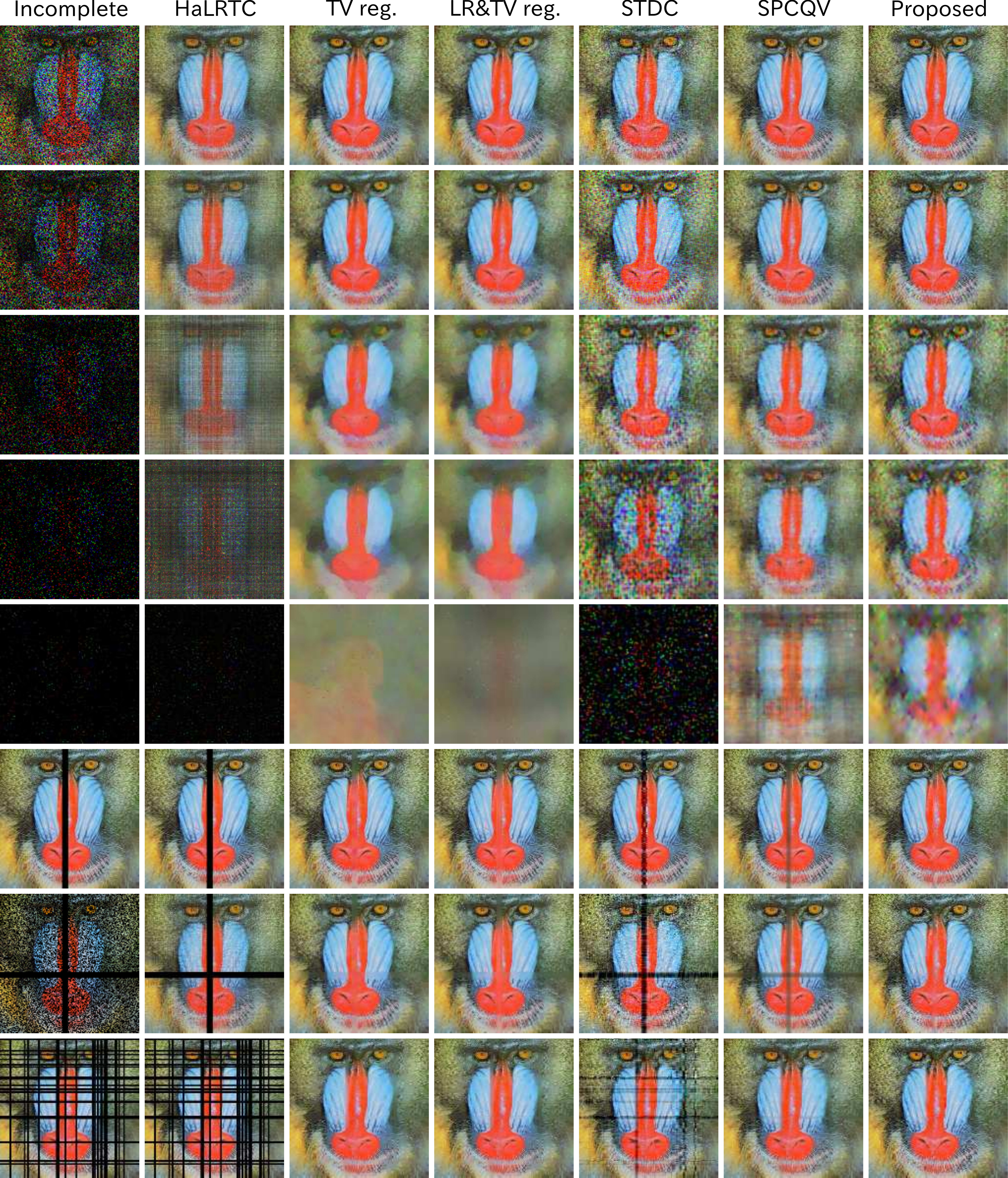}
\caption{Results in `Baboon'}
\end{figure*}

\begin{figure*}[t]
\centering
\includegraphics[width=0.99\textwidth]{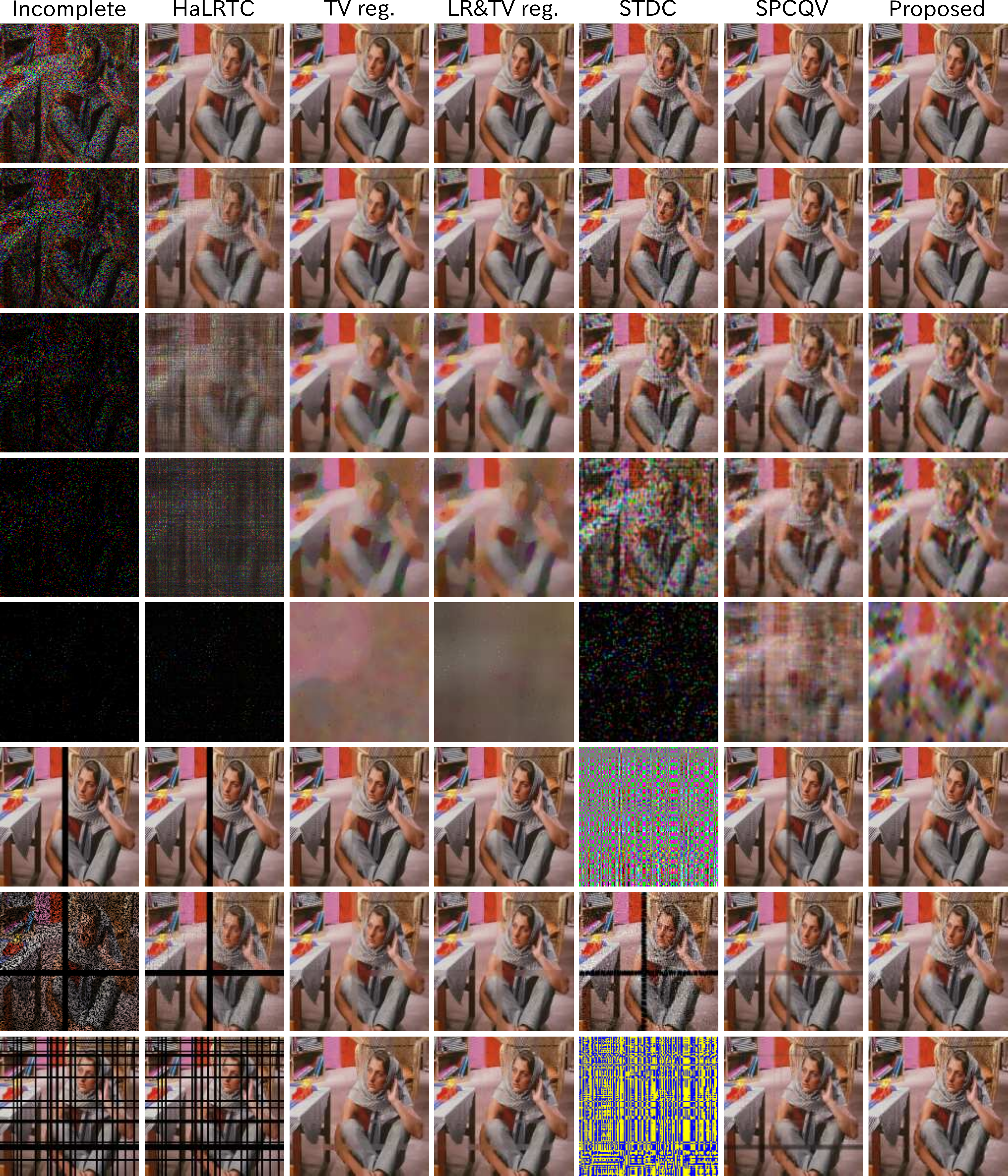}
\caption{Results in `Barbara'}
\end{figure*}

\begin{figure*}[t]
\centering
\includegraphics[width=0.99\textwidth]{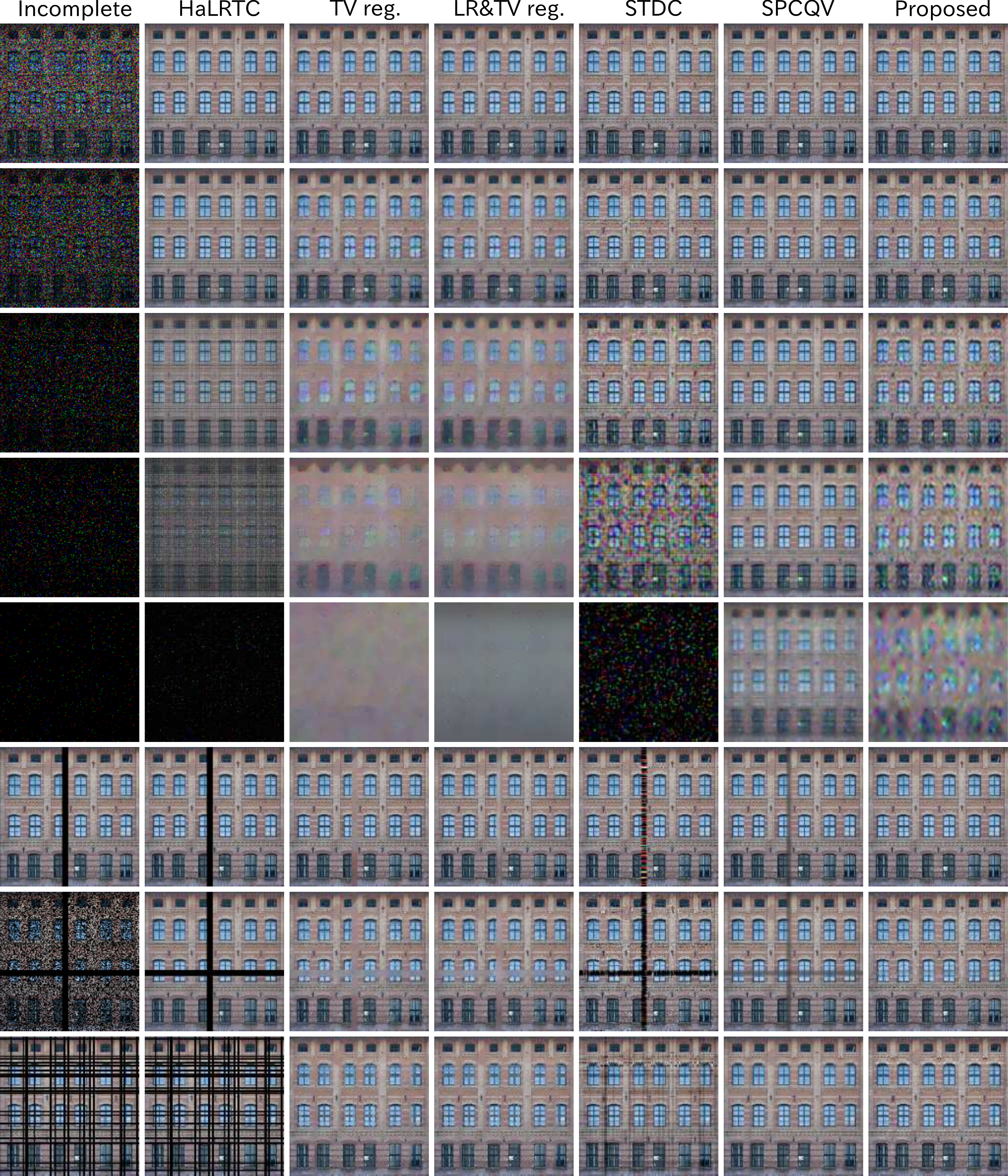}
\caption{Results in `Facade'}
\end{figure*}

\begin{figure*}[t]
\centering
\includegraphics[width=0.99\textwidth]{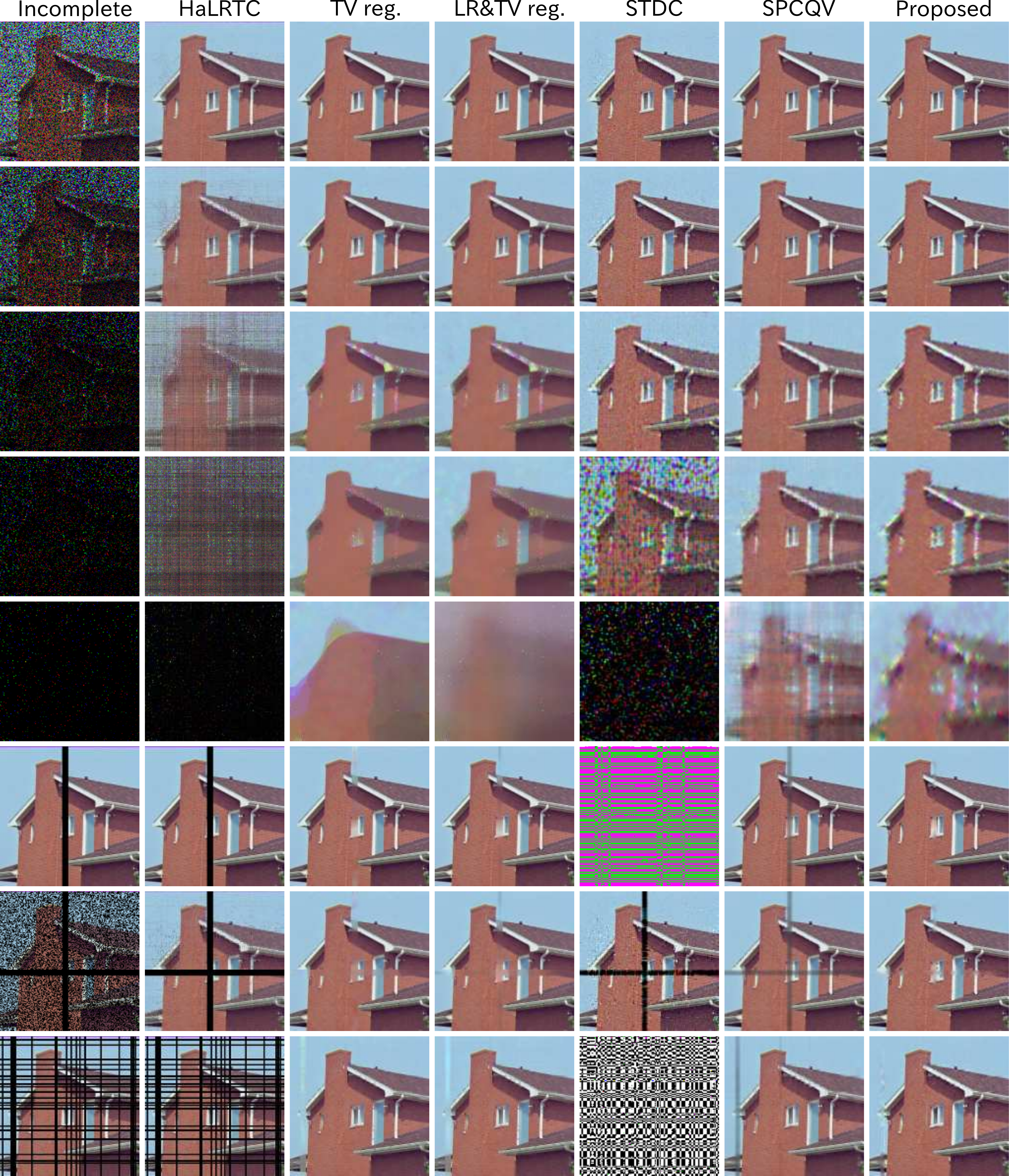}
\caption{Results in `House'}
\end{figure*}

\begin{figure*}[t]
\centering
\includegraphics[width=0.99\textwidth]{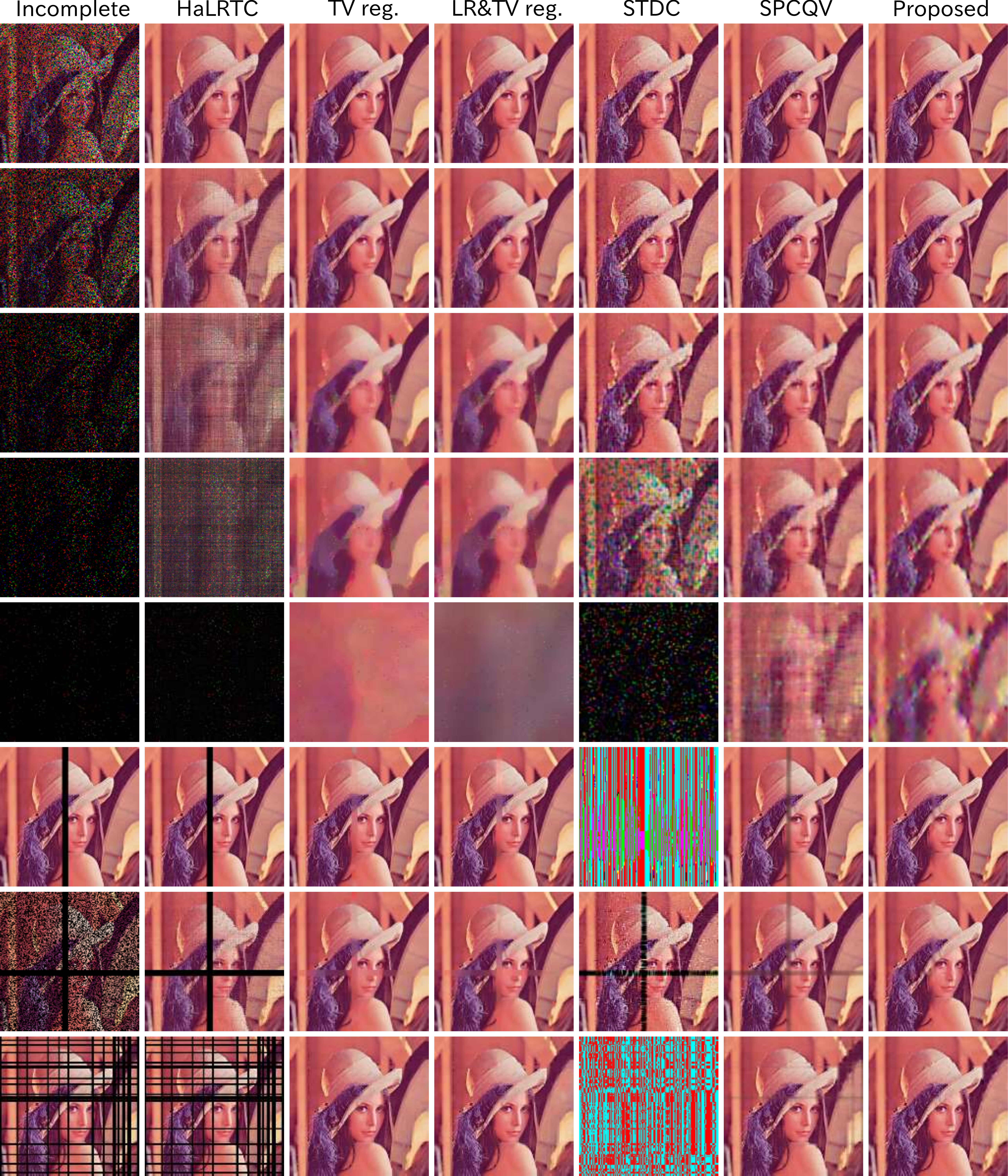}
\caption{Results in `Lena'}
\end{figure*}

\begin{figure*}[t]
\centering
\includegraphics[width=0.99\textwidth]{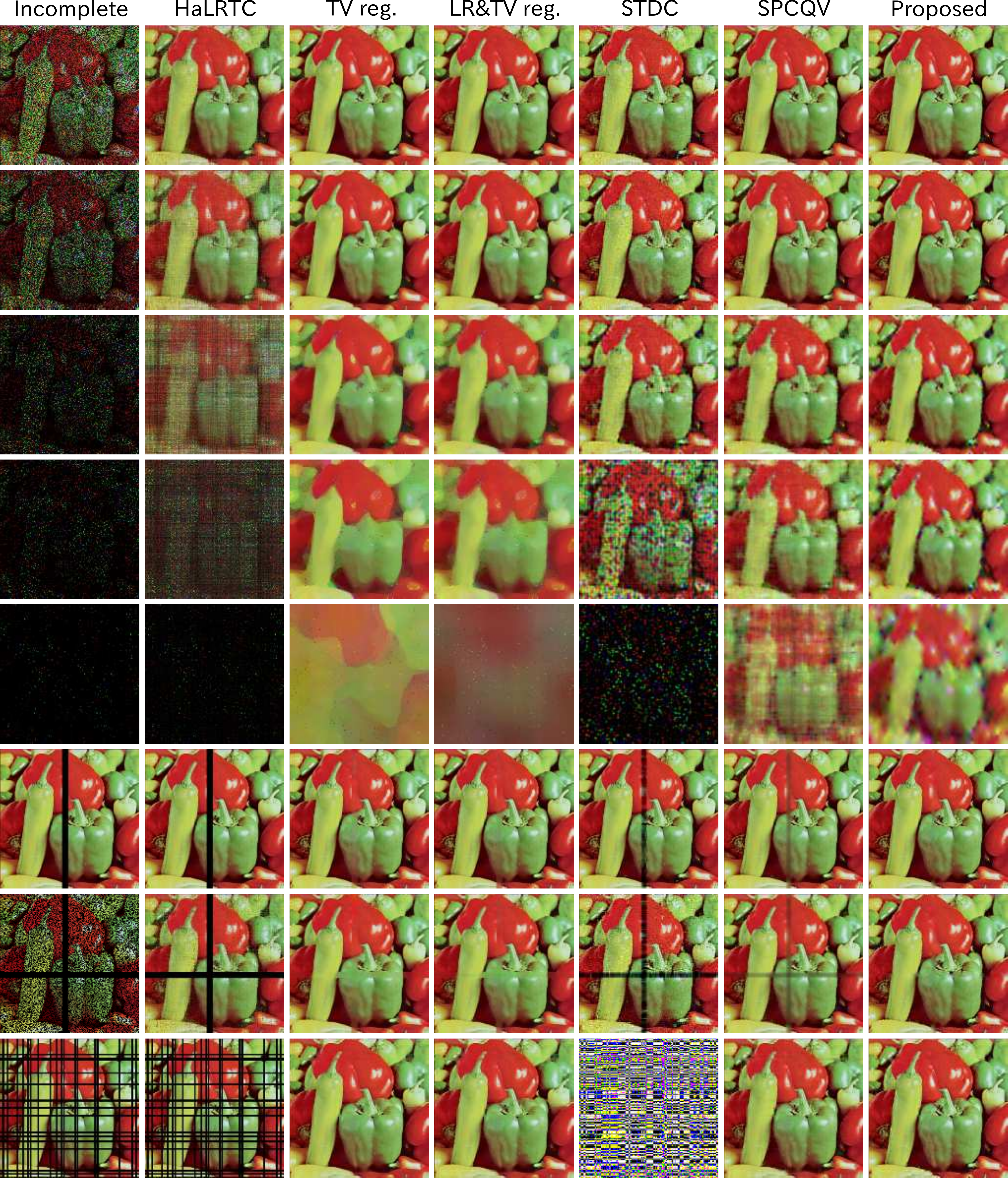}
\caption{Results in `Peppers'}
\end{figure*}

\end{document}